\documentclass{article}

\PassOptionsToPackage{numbers, compress}{natbib}

\usepackage[preprint]{neurips_2021}

\usepackage{microtype}
\usepackage{xcolor}
\usepackage{booktabs} 
\usepackage{url}  
\usepackage{helvet} 
\usepackage{courier}  
\usepackage{graphicx} 
\usepackage{natbib}  
\usepackage{caption} 
\usepackage{subcaption}

\usepackage{amsthm}
\usepackage{color}
\usepackage{comment}
\usepackage{amsmath}
\usepackage{amsthm} 
\usepackage{amssymb}
\usepackage{bbm}
\usepackage{mathbbol} 
\usepackage{soul}
\usepackage{stackengine}
\usepackage{mathtools} 
\usepackage{adjustbox}

\newcommand\barbelow[1]{\stackunder[1.2pt]{$#1$}{\rule{.8ex}{.075ex}}}

\DeclareSymbolFontAlphabet{\mathbb}{AMSb}
\DeclareSymbolFontAlphabet{\mathbbl}{bbold}
\usepackage{wrapfig}

\newcommand{\TV}[2]{\text{TV} [#1,#2]}

\newcommand{\dpiz}{\ensuremath{d_{\pi^z}}}
\newcommand{\dpik}{\ensuremath{d_{\pi^k}}}

\newcommand\defn{\stackrel{\mathclap{\tiny\mbox{def}}}{=}}

\newcommand{\vem}{\ensuremath{v_e^*}}
\newcommand{\vepi}{\ensuremath{v^\pi_e}}

\newcommand{\wa}{\ensuremath{\bar{\w}}}
\newcommand{\wapi}{\ensuremath{\wa^{\text{SMP}}_{\Pi^n}}}

\newcommand{\mat}[1]{\ensuremath{\boldsymbol{\mathrm{#1}}}}
\newcommand{\vphi}{\mat{\phi}}
\newcommand{\vpsi}{\mat{\psi}}
\newcommand{\R}{\ensuremath{\mathbb{R}}}

\newcommand{\w}{\mat{w}}

\newcommand{\ie}{{\sl i.e.}}

\usepackage[utf8]{inputenc} 
\usepackage[T1]{fontenc}    
\usepackage{url}            
\usepackage{booktabs}       
\usepackage{amsfonts}       
\usepackage{nicefrac}       
\usepackage{microtype}      
\usepackage{wrapfig}

\usepackage{natbib}
\usepackage{graphicx}
\usepackage{microtype}
\usepackage{amsmath}
\usepackage{amssymb}

\usepackage{booktabs} 
\usepackage{thmtools}
\usepackage{thm-restate}
\newtheorem{lemma}{Lemma}
\newtheorem{theorem}{Theorem}

\usepackage{hyperref}

\usepackage[capitalize]{cleveref}

\usepackage{algorithm}
\usepackage{algorithmic}

\begin{document}

\author{Tom Zahavy, Brendan O'Donoghue, Andre Barreto, Volodymyr Mnih,\\\textbf{ Sebastian Flennerhag and Satinder Singh}\\ \{tomzahavy,bodonoghue,andrebarreto,vmnih,flennerhag,baveja\}@deepmind.com\\ DeepMind, London}

\title{Discovering Diverse Nearly Optimal Policies with Successor Features}

\maketitle

\begin{abstract}
Finding different solutions to the same problem is a key aspect of intelligence associated with creativity and adaptation to novel situations. In reinforcement learning, a set of diverse policies can be useful for exploration, transfer, hierarchy, and robustness. We propose Diverse Successive Policies, a method for discovering policies that are diverse in the space of Successor Features, while assuring that they are near optimal. We formalize the problem as a Constrained Markov Decision Process (CMDP) where the goal is to find policies that maximize diversity, characterized by an intrinsic diversity reward, while remaining near-optimal with respect to the extrinsic reward of the MDP. We also analyze how recently proposed robustness and discrimination rewards perform and find that they are sensitive to the initialization of the procedure and may converge to sub-optimal solutions. To alleviate this, we propose new explicit diversity rewards that aim to minimize the correlation between the Successor Features of the policies in the set. We compare the different diversity mechanisms in the DeepMind Control Suite and find that the type of explicit diversity we are proposing is important to discover distinct behavior, like for example different locomotion patterns.
\end{abstract}

\section{Introduction}
Creative problem solving is the mental process of searching for an original and previously unknown solution to a problem \citep{osborn1953applied}. The relationship between creativity and intelligence is widely recognized across many fields; for example, in the field of Mathematics, finding different proofs to the same theorem is considered elegant and often leads to new insights. 

Closer to Artificial Intelligence (AI), consider the field of game playing and specifically the game of Chess in which a move is considered creative when it goes beyond known patterns \citep{kasparov}. In some cases, such moves can only be detected by human players while remaining invisible to currently state-of-the-art Chess engines. A famous example thereof is the winning move in game eight of the Classical World Chess Championship 2004 between Leko and Kramnik \citep{chessgame}. Humans and indeed many animals employ similarly creative behavior on a daily basis; faced with a challenging problem we often consider qualitatively different alternative solutions. 

Yet, the majority of AI research is focused on finding a single best solution to a given problem. For example, in the field of Reinforcement Learning (RL), most algorithms are designed to find a single reward-maximizing policy. However, for many problems of interest there may be many qualitatively different optimal or near-optimal policies; finding such diverse set of policies may help an RL agent become more robust to changes in the task and/or environment and to generalize better to future tasks. 

In the field of Quality-Diversity (QD), evolutionary algorithms are used to find useful diverse policies (e.g., \citep{pugh2016quality,mouret2015illuminating,hong2018diversity,masood2019diversity,parker2020effective,gangwani2020harnessing,peng2020non,pmlr-v97-zhang19q}). In a related line of work, intrinsic rewards are used to find diverse skills for fast adaptation \citep{gregor2016variational,eysenbach2018diversity} to be robust to model miss-specification \citep{kumar2020one,zahavy2020planning} and for exploration \citep{agarwal2020pc}. It was also suggested that policies that maximize diversity are more correlated with human behaviour than those that maximize only the extrinsic reward \citep{matusch2020evaluating}. 

This work makes the following contributions. 
\textbf{First,} we propose an incremental method for discovering a diverse set of near-optimal policies. Each policy in the set is trained to solve a Constrained Markov Decision Process (CMDP). The main objective in the CMDP is to maximize the diversity of the growing set, measured in the space of Successor Features \citep[SFs;][]{barreto2017successor}, and the constraint is that the  policies are near-optimal. 
\textbf{Second,} we analyze how previously proposed robustness and discrimination mechanisms for the ``no-reward'' setting perform in terms of diversity in our setup. We find that they are sensitive to the initialization of the procedure and may converge to sub-optimal solutions. To alleviate this, we propose two explicit diversity rewards that aim to minimize the correlation between the SFs of the policies in the set. 
\textbf{Third,}
we demonstrate our method in the DeepMind Control Suite \citep{tassa2018deepmind}. Given an extrinsic reward (e.g. for standing or walking) our method discovers qualitatively diverse locomotion behaviours for approximately maximizing this reward.  

\section{Preliminaries and Notation}
\label{sec:preliminaries}
An MDP \citep{puterman2014markov} is a tuple $M \triangleq (\mathcal{S},\mathcal{A},P, r, \gamma,\rho)$, where $\mathcal{S}$ is the set of states, $\mathcal{A}$ is the set of actions, $P=\{P^a \mid {a \in \mathcal{A}}\}$ is the set of transition kernels, $\gamma \in [0,1)$ is the discount factor and $\rho$ is the initial state distribution. The function $r: \mathcal{S} \times \mathcal{A} \mapsto \R$ defines the rewards. A policy in $M$, denoted by $\pi$, is a mapping $\pi: \mathcal{S} \to \mathcal{P}(\mathcal{A})$, where $\mathcal{P}(\mathcal{A})$ is the probability distributions over $\mathcal{A}$. 

Usually in RL, the agent's objective is to maximize the expected cumulative extrinsic reward. In this work, we will also be interested in discovering and maximizing intrinsic reward functions \citep{singh2010intrinsically}. These rewards can be a function of the policy (e.g., its entropy) or a function of observed features. Let $\vphi(s,a) \in [0,1]^d$ be an observable vector of bounded features. Then there is a set of rewards induced by all possible linear combinations of the features $\vphi$. Specifically, for any $\w \in \R^d$, we can define a reward function $r_{\w}(s,a)=\w \cdot \vphi(s,a).$  Given $\w$, the intrinsic reward $r_{\w}$ is well defined and we will use $\w$ and $r_{\w}$ interchangeably to refer to it. 
Any policy induces a state transition matrix $P^\pi$, where $P^\pi(x,y) = P^{\pi(x)}(x,y)$ is the probability of transitioning from state $x$ to state $y$ when the action is selected according to $\pi(x)$. Thus, any policy yields a Markov chain $(\mathcal{S}, P^\pi)$. By looking at the Markov chain induced by a policy we can study its long-term behavior, such as its stationary distribution. This in turn allows us to define a notion of diversity based on the limiting behavior of policies, in contrast with most previous work on diversity that focus on short-term behavior~\citep{gregor2016variational,eysenbach2018diversity}.

Concretely, in defining diversity we use measures defined in the \textbf{average-case} setting. 
The \textbf{stationary distribution} $d_\pi$ of a Markov chain with transition matrix $P^\pi$ is defined to be $d_\pi(s) = \lim_{t \rightarrow \infty} \Pr (s_t = s |s_0 \sim \rho, \pi)$, which we assume exists and is independent of $s_0$ for all policies. In {\em ergodic MDPs} this limit is unique and is known to be the probability distribution satisfying  $d_\pi^\top=d_\pi^\top P^\pi$
\citep{puterman2014markov}. The asymptotic average reward value, hereafter simply value, of a policy $\pi$ under reward function $r$, denoted $v^\pi_r$, can be defined as an expectation over $d_\pi$ as: $v^\pi_r = \mathbb{E}_{s \sim d_\pi} \, r(s,\pi(s)) = d_\pi \cdot r_\pi,$
where $r_\pi$ is a vector with $\mathbb{E}_{a\sim \pi(s)}r(s,a)$ in its coordinates. 
A natural time scale in this long-term average-case context is the \textbf{mixing time} of the policy -- the time until the Markov chain is "close" to its stationary state distribution. Formally, the $\epsilon$-mixing time $T_\text{mix}$ of an ergodic Markov chain with a stationary distribution $d_\pi$ is the smallest time $t$ such that $\forall x_0,\ \TV{\Pr_t(\cdot|x_0)}{d_\pi} \le \epsilon$, where $\Pr_t(\cdot|x_0)$ is the distribution over states after $t$ steps, starting from $x_0$ and $\TV{\cdot}{\cdot}$ is the total variation distance. In other words, if we follow a policy in an MDP for $T_{\text{mix}}$ steps, we will observe states that are approximately distributed according to $d_\pi$. 


Similarly, we can define the expected features, also known as \textbf{successor features}, under $d_\pi$ as $\psi^\pi = \mathbb{E}_{x \sim d_\pi} \, \phi(x,\pi(x)).$
Note that the SFs are conditioned on $\rho$ and $\pi$ and that they are vectors in $\mathbb{R}^d$; similar definitions were suggested in \citep{mehta2008transfer,zahavy2019average}. For linear rewards there is a simple way to express the average reward value of the policy (\cref{sec:preliminaries}) using the SFs: 
$
v^\pi_w = \psi^\pi\cdot w.
$
To keep the notation simple, we will refer to the SFs of policy $\pi^i$ as $\psi^i$; and, since we are dealing with different intrinsic rewards, we will use the notation $v_d^i$ to refer to the value of policy $\pi^i$ for reward $r_d$.

\section{Discovering diverse near-optimal policies}
\label{sec:problem}
We are interested in discovering a set of $n$ near-optimal policies $\Pi^n= \{ \pi^i \}_{i=1}^n$ that are maximally diverse according to some diversity metric. 
Let $\Psi^n$ be the set of SFs corresponding to the policies in $\Pi^n$, then we are interested in solving the following constrained optimization problem:
\begin{equation}
    \label{eq:problem_def}
    \max_{\Pi^n}  \enspace \text{Diversity}(\Psi^n) \enspace
    \text{s.t} \enspace \enspace \vepi \ge \alpha \vem, \enspace \forall \pi \in \Pi^n,
\end{equation}
where $\text{Diversity}: \{ \mathbb{R}^d \}^n \rightarrow \R$ measures the diversity of a set of SFs ($\Psi^n$) that we shall define shortly, and the constraint requires that all the policies in $\Pi^n$ achieve value better than a parameter $\alpha \in [0, 1]$ times the value of the optimal policy (here $\vepi$ is the value of policy $\pi$ for extrinsic reward $r_e$ and $\vem$ is the value of the optimal policy with respect to $r_e$). Note that $\alpha$ controls how big a space of policies we search over for our diverse set of policies. In general, the smaller the $\alpha$ parameter the larger the set of $\alpha$-optimal policies and thus the greater the diversity of the policies found in $\Pi^n$\footnote{When the extrinsic reward is positive ($r_e(s,a)\ge0, \forall s,a$), the extrinsic value is positive $v_e^\pi\ge0, \forall \pi,$ and setting $\alpha=0$ in \cref{eq:problem_def} is equivalent to the no-reward setting where the goal is to maximize diversity.}.

\begin{wrapfigure}{r}{0.6\textwidth}
\begin{minipage}{0.6\textwidth}
\vspace{-0.8cm}
\begin{algorithm}[H]
\caption{Diverse Successive Policies}\label{alg:qd}
\begin{algorithmic}[1]
\STATE \textbf{Input:} mechanism to compute rewards $r_e$ and $r_d.$
\STATE \textbf{Initialize:} $\pi^0 \leftarrow \arg\max_{\pi \in \Pi} r_e \cdot d_\pi$, 
\STATE $\vem = v^{\pi^0},$ $\Pi^0 = \{\pi^0\}$
\FOR{$i = 1,\ldots,T$}
    \STATE Compute diversity reward $r^i_d = \text{D}(\Psi^{i-1})$
    \STATE $
    \pi^i = \arg\max _{\pi}   d_{\pi} \cdot r^i_d  \enspace \enspace
    \text{s.t.} \enspace  \enspace d_{\pi} \cdot r_e \ge \alpha \vem
    $
    \STATE Estimate the SFs $\psi^i$ of the policy $\pi^i$
    \STATE $\Pi^i = \Pi^{i-1} \cup \{\pi^i\}, \Psi^i = \Psi^{i-1} \cup \{\psi^i\}$
\ENDFOR
\STATE \textbf{return} $\Pi^{T}$
\end{algorithmic}
\end{algorithm}
\vspace{-0.5cm}
\end{minipage}
\end{wrapfigure} 
Common to many approaches is to define a diversity objective using \emph{intrinsic} rewards \citep{gregor2016variational,eysenbach2018diversity,kumar2020one,zahavy2021discovering}, \ie, rewards not from the environment but defined by the agent itself. Our approach also uses intrinsic rewards to induce diversity, as we describe in \cref{alg:qd}. The algorithm receives as input two reward functions $r_e$ and $r_d$, which together define a CMDP. The reward $r_d$ corresponds to a diversity intrinsic reward. We will discuss five different candidate $r_d$'s. The constraint reward $r_e$ will typically be the extrinsic reward, but we will also consider two alternative choices for $r_e$. 
In the initialization step of \cref{alg:qd} (line 2) there are no policies in the set, and so the goal of the first policy $\pi^0$ is to solve the MDP with reward $r_e$. \cref{alg:qd} then adds $\pi^0$ and its SFs to the set, and the variable $\vem$ is set to be $v^0.$ $\vem$ defines the near-optimality constraint $\alpha \vem$ for the other policies (say with $\alpha=0.9$). 

After this first step, the algorithm proceeds in iterations. In iteration $i$, an intrinsic reward $r^i_d$ is computed given the previous policies in the set $\Pi^{i-1}$.  The next policy to be added to the set, $\pi^i$, is the solution to the following {Constrained MDP (CMDP)} (line 6 in \cref{alg:qd}):  
\begin{equation}
    \label{eq:cmdp}
    \arg\max_{\pi} \enspace d_{\pi} \cdot r^i_d  \enspace  \enspace
    \text{s.t.}  \enspace \enspace  d_{\pi} \cdot r_e \ge \alpha \vem.
\end{equation}
In words, the new policy optimizes the average intrinsic reward value subject to the constraint that it be near-optimal with respect to its average extrinsic reward value. In \cref{sec:solvecmdp} we discuss the details of how to solve \cref{eq:cmdp}. 
Clearly, the behavior of \cref{alg:qd} strongly depends on the choice of $r_d$, the intrinsic reward used to induce diversity. We now discuss five alternatives to define this reward.
\section{Measuring Policy Diversity}
\label{sec:diversity}

A key aspect of our method is the measure of diversity. Our focus is on diverse \emph{policies}, as measured by their stationary distribution after they have mixed. This suggests we should measure diversity in the space of SFs, as they are defined under the policy's stationary distribution (see \cref{sec:preliminaries}). In contrast, prior work have focused on learning diverse \emph{skills}, which is often measured before the skill policy mixes. A common approach to measuring skill diversity is to measure skill discrimination in terms of trajectory-specific quantities such as terminal states \citep{gregor2016variational}, a mixture of the initial and terminal states \citep{baumli2020relative}, or trajectories \citep{eysenbach2018diversity}. An alternative approach that implicitly induces diversity is to learn policies that maximize the robustness of the set $\Pi^n$ to the worst-possible reward \citep{kumar2020one,zahavy2021discovering}.

In Subsections \ref{subsec:discrimination} and \ref{subsec:robustness}, we analyze the diversity of these two approaches in the space of SFs and find that they both depend on the initialization of the algorithm and cannot guarantee diversity. Motivated by these findings, we develop two new explicit diversity rewards that aim to minimize the correlation between the SFs of the policies in the set. We discuss these new methods in \cref{subsec:explicit}. 
\subsection{Diversity via Discrimination}
\label{subsec:discrimination}
Discriminative approaches rely on the intuition that skills should be distinguishable from one another simply by observing the states that they visit. Learning diverse skills is then a matter of learning skills that can be easily discriminated. For instance, DIAYN \citep{eysenbach2018diversity} maximizes the mutual information between skills and states as follows. Given a probability space $(\Omega, \mathcal{F}, \mathcal{P})$, we denote by $I(S;Z)$ the mutual information between the random variable state $S: \Omega \rightarrow \mathcal{S}$ and latent random variable (skill) $Z : \Omega \rightarrow \mathcal{Z}$ \citep{cover1999elements}. We also use $H[A | S]$ to refer to the conditional entropy of the action random variable $A: \Omega \rightarrow \mathcal{A}$ conditioned on state $S$. Finally, the conditional mutual information between $A$ and $Z$ given $S$ is denoted by $I(A;Z | S)$. Then, the DIAYN objective to be maximized, given a prior over the latents, $p$, is:
\begin{align}\label{eq:diyan}
I(S;Z) + H[A | S] - I(A;Z | S) = H[A|S,Z] + \mathbb{E}_{\substack{z \sim p(z)\\ s \sim \dpiz}} [\log p (z | s)-\log p(z)].
\end{align}
This is an entropy-regularized objective that seeks to maximize the information that states contain about the skill used to reach it. In particular, the term of interest is $\mathbb{E}_{\substack{z \sim p(z), s \sim \dpiz}} [\log p (z | s)-\log p(z)]$, which corresponds to the value of a skill in an MDP with reward $r(s | z) = \log p (z | s) - \log p(z)$. A skill policy $\pi(a | s, z)$ controls the first component of this reward, $p(z|s)$, which measures the probability of identifying the skill in state $s$. Hence, the policy is rewarded for visiting states that differentiates it from other skills, thereby implicitly encouraging diversity.

The exact form of $p(z | s)$ depends on how skills are encoded \citep{gregor2016variational}. The most common version is to encode $z$ as a one-hot $d$-dimensional variable \citep[e.g.;][]{gregor2016variational,achiam2018variational,eysenbach2018diversity}. Similarly, we represent $z$ as $z \in \left\{1, \ldots, n \right\}$ to index $n$ separate policies $\pi^z$. In addition, the concept of finding a small set of meaningful policies is appealing from the interpretability perspective.  

$p(z|s)$ is typically intractable to compute due to the large state space and is instead approximated via a learned discriminator $q_\phi(z | s)$. In our case, we measure $p(z | s)$ under the stationary distribution of the policy; that is, $p(s|z) = \dpiz(s)$. Therefore, for the purpose of analysis, we can find an analytic form for the objective of DIAYN before we apply the variational approximation. Given this, applying Bayes rule to $p(z | s)$ yields
\begin{equation}
\label{eq:bayes}
    p(z|s) = \frac{\dpiz(s)p(z)}{\sum_{k} \dpik(s)p(k)}.
\end{equation}
And in the kernel case, we define a Gibbs distribution
\begin{equation}
\label{eq:bayes_sfs}
    p(z|s) = \frac{p(z)\exp{(\phi(s) \cdot \psi^z)}}{\sum p(k)\exp(\phi(s) \cdot \psi^k)}.
\end{equation}
Plugging $p(z|s)$ from \cref{eq:bayes} in the objective of DIAYN, the relevant term in \cref{eq:diyan} becomes
\begin{align}
    \mathbb{E}_{z \sim p(z), s\sim d(\pi^z)} [\log p (z | s)] 
    = \sum_z p(z) \sum_{s} \dpiz(s)  \log \left( \frac{\dpiz(s)p(z)}{\sum_{k} \dpik(s)p(k)}\right).    \label{eq:pivic}
\end{align}
Finding a policy with maximal value for this reward can be seen as solving an optimization program in $\dpiz$ under the constraint that the solution is a valid stationary state distribution (\cref{sec:preliminaries}). The term $\sum_s p(s | z) \log p(s | z)$ corresponds to the negative entropy of $\dpiz$, meaning that the objective to be maximized is \emph{convex} in $\dpiz$. 
\begin{lemma}
\label{lemma:conex_vic}
The function $\sum_{s} \dpiz(s)  \log \left( \frac{\dpiz(s)p(z)}{\sum_{k} \dpik(s)p(k)}\right)$ is a convex function of  $\dpiz.$
\end{lemma}
The proof can be found in \cref{sec:proofvic}; briefly, \cref{lemma:conex_vic} holds because the function can be written as $KL(\dpiz || \sum_{k} p(k) \dpik) + \sum_{s} \dpiz(s) \log p(z)$ and the KL-divergence is jointly convex on both arguments \citep[Example 3.19]{boyd2004convex}.  
%
The convexity of the objective results from the fact that the intrinsic reward $\log p(z|s)$ is a function of the policy. In the standard RL setup, the reward is not a function of the policy and the objective is linear in it, thus, maximizing and minimizing the reward are both convex minimization problems. However, when the reward is a function of the policy, maximization and minimization of the reward are not equivalent optimization problems. In  DIAYN, the maximization of $\log p(z|s)$ leads to convex maximization while the minimization of the same reward leads to convex minimization. We note that the convexity of the objective has nothing to do with the variational approximation typically used to compute $p(z|s)$; it is encountered with or without it.

The observation that discriminatory objectives lead to a set of $n$ convex maximization problems in our setting is problematic, since the optimality of the solutions---in particular, their diversity---cannot be guaranteed. From the perspective of the policy set, the algorithm may converge to a set which is a local maxima rather than the global maxima, and therefore result in suboptimal diversity. In practice, different initializations and stochastic updates might mitigate the issue to some degree. In addition, it is possible that all the local maxima are close to optimal. For example, similar observations were made regarding the loss surface of deep neural networks, but the local optima points were shown to be very good in practice \cite{NIPS2014_17e23e50,choromanska2015loss,soudry2016no}, mitigating the issues mentioned above. Thus, we recommend taking \cref{lemma:conex_vic} as an observation regarding the optimization landscape of DIAYN which we hope to further explore in future work. 


\subsection{Diversity via Robustness}
\label{subsec:robustness}

An alternative approach that implicitly induces diversity is to seek robustness among a set of policies by maximizing the performance w.r.t the worst case reward \citep{kumar2020one,zahavy2021discovering}; for fixed $n$, the goal is:
\begin{equation}
\label{eq:max_min_max}
    \max _{\Pi^n \subseteq \Pi} \min_{w\in B_2} \max_{\pi^i \in \Pi^n} \psi^i \cdot w.
\end{equation}
Here $B_2$ is the $\ell_2$ unit ball, $\Pi$ is the set of all possible policies, $\Pi^n = \{\pi^1, \ldots, \pi^n\}$ is the set of $n$ policies for which we are optimizing. Let us parse this objective term by term. First, the inner product $\psi^i \cdot w$ yields the expected value under the steady-state distribution (see \cref{sec:preliminaries}) of the policy $\pi^i$. The inner min-max is a two-player zero-sum game, where the minimizing player is finding the \emph{worst-case} reward function (since weights and reward functions are in a one-to-one correspondence) that minimizes the expected value, and the maximizing player is finding the best policy from the set $\Pi^n$ (since policies and SFs are in a one-to-one correspondence) to maximize the value. The outer maximization is to find the best set of $n$ policies that the maximizing player can use.

Intuitively speaking, the solution $\Pi^n$ to this problem might be a diverse set of policies
since a non-diverse set is likely to yield a low value of the game, that is, it would easily be exploited by the minimizing player. In this way diversity and robustness are dual to each other, in the same way as a diverse financial portfolio is more robust to risk than a heavily concentrated one. By forcing our policy set to be robust to an adversarially chosen reward it will be diverse. 

In \citep{kumar2020one}, the authors proposed a solution to \cref{eq:max_min_max} using a CMDP with $r_d$ as discrimination (via DIAYN) and $r_e$ is the extrinsic reward; we discuss it in more detail in \cref{sec:solvecmdp}. In \citep{zahavy2021discovering}, the authors proposed an iterative solution to \cref{eq:max_min_max} that incrementally adds policies to a solution set $\Pi^n$ (\cref{alg:greedy-iterative} in the appendix). The authors define a Set Max Policy (SMP) as a policy that takes a set of policies and a reward as inputs and returns the best policy in the set for this reward. In each iteration, the algorithm computes the worst case reward w.r.t to the SMP, finds the policy that maximizes it, and adds it to the set. In iteration $n$ The value of the SMP on the set $\Pi^n$ is defined as $v^n = \min_{w\in B_2} \max_{\pi^i \in \Pi^n} \psi^i \cdot w$, and it is guaranteed that this value strictly increases $v^{n+1}>v^n$ in each iteration until the optimal solution is found.
The following Lemma suggests that this procedure is equivalent to a fully corrective FW \citep{frank1956algorithm} algorithm on the function $f=||\cdot||_2.$ As a consequence, it is guaranteed to convergence to the optimal solution in a linear rate \citep{jaggi2015global}. 
\begin{lemma}
\label{FW:wcpi}
The iterative procedure in \cite{zahavy2021discovering} is equivalent to a fully corrective FW algorithm to minimize the function $f=||\psi^\pi||_2.$ As a consequence, to achieve an $\epsilon-$optimal solution, the algorithm requires at most $O(\log(1/\epsilon))$ iterations. 
\end{lemma}
The proof in \cref{sec:wcpi} suggests that the SMP policy is equivalent to the fully corrective search (maintaining a dictionary of solutions from previous iterations and choosing the best convex combination). The only difference between the two algorithms is that one of them solves a max-min problem where the other solves the equivalent min-max problem, and therefore they are guaranteed to have the same iterations from strong duality. 
Unfortunately this approach, like the discriminative approaches, has a weakness that can limit the ultimate diversity in the set. To see this note that
\begin{align*}
&\max _{\Pi^n \subseteq \Pi} \min_{w\in B_2} \max_{\pi^i \in \Pi^n} \psi^i \cdot w 
\leq \min_{w\in B_2}\max _{\pi^i \in \Pi}  \psi^i \cdot w
= \max _{\pi^i \in \Pi}\min_{w\in B_2}  \psi^i \cdot w
= - \min_{\pi^i \in \Pi} \|\psi^i\| \defn v^*,
\end{align*}
where the inequality comes from the fact that $\Pi^n \subseteq \Pi$, and the first equality uses von Neumann's minimax theorem \citep{neumann1928theorie}.
If we let $
    \pi^{*} = \arg \min_{\pi^i \in \Pi} \|\psi^i\|,$ then if $\Pi^n = \{\pi^*\}$ we have an optimal
policy set for the game, since we have found a policy set that achieves the known upper bound on the value of the game, $v^*$. In other words a single policy is a sufficient solution for  \cref{eq:max_min_max}, which is problematic since the goal was to build up a set of many diverse policies. Similar to the discriminative approaches, in practice we obtain more policies by initializing the set away from $\pi^*$, or alternatively restricting $\Pi^n$ to deterministic policies. However, this issue likely explains the empirical observations in \cite{zahavy2021discovering} that there are only a few active policies in the optimal sets.

Note that the results above hold only in the case that $\Pi$ is the set of all the stochastic policies in the MDP; if only deterministic policies are used, we cannot apply the von Neumann's minimax theorem. 
This is not an issue since we are interested in stochastic policies for multiple reasons: optimal solutions to CMDPs are stochastic policies \citep{altman1999constrained} and stochastic policies are the most common approach in continuous control tasks, which is the focus of our experiments. 

\subsection{Explicit diversity methods}
\label{subsec:explicit}

The two diversity mechanisms we have discussed so far were designed to maximize robustness or discrimination. Each one has its own merits in terms of diversity, but since they do not explicitly maximize a diversity measure they cannot guarantee that the resulting set of policies will be diverse. 
%
%
We now propose two reward signals designed to induce a diverse set of policies. The way they do so is to leverage the information about the policies' long-term behavior available in their SFs. Both rewards are based on the intuition that the correlation between SFs should be minimized.

To motivate this approach, we note that SFs can be seen as a compact representation of a policy's stationary distribution. This becomes clear when we consider the case of a finite MDP with $|\mathcal{S}|$-dimensional ``one-hot'' feature vectors $\phi$ whose elements encode the states: $\phi_i(s) = \mathbb{I}\{ s = i \}$, where $\mathbb{I}\{ \ \cdot \}$ is the indicator function. In this special case the SFs of a policy $\pi$ coincide with its stationary distribution, that is, $\psi^{\pi} = d_{\pi}$. Under this interpretation, minimizing the correlation between SFs intuitively corresponds to encouraging the associated policies to visit different regions of the state space---which in turn leads to diverse behavior. As long as we assume the tasks of interest are linear combinations of the features $\phi \in \R^d$, which we do, similar reasoning applies when $d < |\mathcal{S}|$. 

But how do we compute policies in order to minimize the correlation between their SFs? To answer this question, we first consider the extreme scenario where there is a single policy $\pi^k$ in the set $\Pi$. In this case the objective is: $\max_{\psi^z} \psi^z \cdot w,$ where $w= -\psi^k.$ Solving this problem is an RL problem whose reward is linear in the features weighted by $w$. A similar objective was investigated in \citep{hansen2019fast}, but there $w$ was sampled i.i.d from a fixed prior. The question we are trying to address is: how to define $w$ taking into account multiple policies in the set $\Pi^n$?

We propose two answers to this question. The first one is to have $w$ be the negative average of the SFs of the policies currently in the set, that is,  $w = - \frac{1}{k} \sum \psi^k$. This formulation is useful as it measures the sum of negative correlations within the set. However, when two policies in the set happen to have the same SFs with opposite signs, they cancel each other, and do not impact the diversity measure. This diversity objective shares some similarities with the novelty search algorithm in \citep{conti2018improving}, where the mean pairwise distance between the current policy and an archive of other policies is used.

The second diversity-inducing reward we propose addresses this issue. It is defined as the minimum over the SFs in each state: $r(s) = \min_k \left\{\phi(s) \cdot -\psi^k\right\}.$ This objective encourages the policy to have the largest "margin" from the set, as it maximizes the negative correlation from the element that is "closest" to it. This objectives shares some similarities with a recent work \citep{parker2020effective} that uses the determinant of the kernel matrix and penalizes it to the closest agents in the population, building on ideas from Determinantal point processes \citep{kulesza2012determinantal}. Finally, we note that we also apply a non linear transformation to bound both of these rewards; the details are in the supplementary (\cref{sec:tech_details}). 

\section{Solving the constrained MDP}
\label{sec:solvecmdp}
At the core of our approach is the solution of a CMDP. The literature on CMDPs is quite vast and we refer the reader to \citep{altman1999constrained}  and \citep{cmdpblog} for treatments of the subject at different levels of abstraction. In this work we will focus on a reduction of CMDPs to MDPs via gradient updates. The idea is to look at the Lagrangian of \cref{eq:cmdp}: 
\begin{equation}
    \label{eq:cmdp_L}
    L(\pi,\lambda) =  - d_{\pi} \cdot (r_d + \lambda r_e) - \lambda \alpha \vem. 
\end{equation}
Then, solving the CMDP in \cref{eq:cmdp} is equivalent to solving 
$
\min_{\pi \in \Pi} \max_{\lambda\geq 0} L(\pi,\lambda).$
%

Solving CMDPs via Lagrangian methods dates back to \cite{borkar2005actor,bhatnagar2012online}; more recently the problem has been tackled using Deep RL techniques  \citep{tessler2018reward,calian2020balancing}. These algorithms perform primal-dual gradient updates on the min-max game. 
When the value function of the policy satisfies the constraint, the Lagrange multiplier will decrease, putting more emphasis on the extrinsic reward; when the constraint is not satisfied, the Lagrange multiplier will increase to satisfy the constraint.

\textbf{Non linear Lagrange multiplier.} We would like our agent to optimize a bounded reward signal, and we discuss how to bound each reward $r_d$ in the supplementary (\cref{sec:tech_details}). To guarantee that a combination of two bounded rewards remains bounded, it is sufficient to combine them via a convex combination. To achieve that, we use a Sigmoid activation on the Lagrange multiplier so the reward is a convex combination of the diversity and the extrinsic rewards: 
$$r(s) = \sigma(\lambda) r_e(s) + (1-\sigma(\lambda)) r_d(s).$$
We further introduce an entropy regularization on $\lambda$ to prevent $\sigma(\lambda)$ from getting to extreme values ($1$ or $0$), where the Sigmoid activation is saturated and has low  gradients. This can happen, for example, at the beginning of learning where the agent's policy is sub-optimal and does not satisfy the constraint for many iterations. 
The objective for $\lambda$ is thus:
\begin{equation}
\label{eq:lagrange}
 f(\lambda) =  \sigma(\lambda) (\hat v - \alpha \vem) - a_h H(\sigma(\lambda)),   
\end{equation}
where $H$ is the entropy function, $a_h$ is the weight of the entropy regularization and $\hat v$ is an estimate of the total cumulative extrinsic return that the agent obtained in recent trajectories. The Lagrangian $\lambda$ is updated by performing gradient descent on \cref{eq:lagrange} every $N_{\lambda}$ agent steps.

\textbf{Estimation of average rewards.}
Another important step of \cref{alg:qd} which is not directly related to solving the CMDP is the estimation of the average rewards. For that, we used a simple Monte Carlo estimates:  $\tilde v_j = \frac{1}{T} \sum_{t=1}^T r_t,$ i.e, the empirical average reward obtained by the agent in trajectory $j$ (where $T=1000$). We used the same estimator to estimate the average SFs (replace $r_t$ with $\phi_t$). 

The value $\tilde v_j$ is a good estimate of the average reward, but it is not perfect. The issue is that the trajectory is of finite length, and therefore the samples in the beginning of the trajectory, before the policy is mixed, are biased. Our experiments are in the DM control suite \citep{tassa2018deepmind} where the mixing time is small; the policies we discover roughly mix after $\sim 50$ steps (as can be seen in the videos in the supplementary). Since the mixing time is much shorter than $T$, the effect of the biased samples is small ($\sim 5\%$). It is also possible to wait until the policy is mixed or to collect a perfect unbiased estimate of the average reward via Coupling From the Past procedure \citep{propp1996exact} as was done in \citep{zahavy2019average}. Note that this is a known issue with any practical policy gradient method but was not found to make a big difference empirically. 

We further average the estimate using a running average with decay factor of $a_d$: $\hat v_j = a_d \hat v_{j-1} + (1-a_d) \tilde v_j$; this is the estimate we use in \cref{eq:lagrange}. The running average variables are set to $0$ between iterations of \cref{alg:qd}.
Finally, we note here that we also experimented with the discounted criteria (discounted SFs). In that case, we observed that there is too much emphasis on the features that are observed at the beginning of the trajectory, resulting in less diversity across the entire trajectory.




\textbf{Discussion.} 
A different feasible approach to combine $r_d$ and $r_e$ is to model the problem as a \textit{multi-objective} MDP. That is, the diversity objective is added to the main one via a fixed, stationary weighting of the two rewards, e.g., $r = a_1 r_d + a_2 r_e$.
We note that the solution of such a multi-objective MDP cannot be be a solution to a CMDP. I.e., it is not possible to find the optimal dual variables $\lambda^*,$ plug them in \cref{eq:cmdp_L} and simply solve the resulting (unconstrained) MDP. Such an approach ignores the fact the dual variables must be a `best-response' to the policy and is referred to as the "scalarization fallacy" in \citep[Section 4]{cmdpblog}.

While Multi objective MDPs have been used in prior QD-RL papers \citep{hong2018diversity,masood2019diversity,parker2020effective,gangwani2020harnessing,peng2020non,pmlr-v97-zhang19q}, we now outline a few potential advantages for using CMDPs. First, the CMDP formulation guarantees that the policies that we find are near optimal (satisfy the constraint). Secondly, the weighting coefficient in multi-objective MDPs has to be tuned, while in our case it is being adapted over time. This is particularly important in the context of maximizing diversity while satisficing reward. In many cases, as we observed in our experiments, the diversity reward might have no other option other than being the negative of the extrinsic reward. In these cases our algorithm will return good policies that are not diverse, while a solution to multi-objective MDP might fluctuate between the two objectives and not be useful at all. 

\textbf{CMDPs in related QD papers.} Kumar et al. \cite{kumar2020one} proposed that solving a CMDP with $r_d$ as discrimination reward and $r_e$ as the extrinsic reward will lead to a solution to the robustness objective (\cref{eq:max_min_max}). Sun et al. \cite{sun2020novel} also investigated CMDPs, but focused on the setup where the diversity reward has to satisfy a constraint, so the diversity reward is $r_e$ and the extrinsic reward is $r_d$. But most importantly, we use a different method to solve CMDPs, which is based on Lagrange multipliers and SFs and is justified from CMDP theory \citep{altman1999constrained,borkar2005actor,bhatnagar2012online}, while these other two papers use techniques that are not guaranteed to solve CMDPs.


\section{Experiments}

We conducted our experiments on domains from the DM Control Suite \citep{tassa2018deepmind}, standard continuous control locomotion tasks where diverse near-optimal policies should naturally correspond to different gaits. We focused on the setup where the agent is learning from feature observations corresponding to the positions and velocities of the body joints being controlled by the agent. Due to space considerations, we focus on domains where the diversity is interesting from a visual point of view, and in particular on Walker and Dog. In simpler domains like Cartpole and Reacher, we observed simple symmetric diversity -- one policy moves a certain way clockwise and then the second policy moves in the same way anti-clockwise (see~\cref{fig:cartpole} in the supplementary). Later policies in the set are less distinguishable visually but can learn, for example, to balance the pole while moving. Note that without a diversity mechanism, the agent tends to only move in a single direction (e.g. clockwise).



In most of our experiments, the extrinsic reward $r_e,$ which defines the optimality constraint in Algorithm \ref{alg:qd}, is set to be the environment reward provided by the DM Control Suite . The first policy in the set is trained to only maximize the extrinsic reward, and the other policies has to satisfy the constraint of being $\alpha = 0.9$ optimal w.r.t it. In these experiments, we report the reward that each policy collects in white color on top of each figure. Additionally we report the reward of each policy in a small table in the main text. 

In the QD community, there is no consensus regarding a single metric for measuring diversity, and some argue that there shouldn't be such (see, for example, the book ”Why Greatness Cannot Be Planned” \citep{stanley2015greatness}). Inspired by this literature, we focus on measuring diversity only qualitatively by visualizing the learned policies. We strongly recommend the reader to check our visualization website where we show \textbf{videos} of the trajectories that each policy takes at \url{https://anon98723.github.io/}. In addition, we present "motion figures" by discretizing the videos (details in the Appendix) that give a fair impression of the policy behaviours. We would like to note that we did not tune our method to maximize diversity based on any metric other than constraint satisfaction (maintaining near-optimality). The main purpose of our experiments are the feasibility of the CMDP framework as proposed in Algorithm \ref{alg:qd}, i.e., to demonstrate that we discover diverse near-optimal policies.   


\textbf{Choice of $r_d$:} Given that our Diverse Successive Policies algorithm (\ref{alg:qd}) can be used with different measures of diversity, we compared four different choices. The previously proposed robustness and discrimination measures and the new min and average explicit measures of diversity we proposed in \cref{subsec:explicit}, corresponding to: \textbf{(1) Robustness:} the worst case linear reward with respect to the previous policies in the set: $r_d(s) = w \cdot \phi(s),$ where $w = \min_{w\in B_2} \max_{z \in [1,..,n-1]} \psi^z \cdot w$ is the internal minimization in \cref{eq:max_min_max}. \textbf{(2) Discrimination} $r_d(s) = \log(\frac{\exp{(\phi(s) \cdot \psi^n)}}{\sum_{i=j}^n \exp(\phi(s) \cdot \psi^j)}),$ where $\psi^n$ is the running average estimator of the SFs of the current policy. This reward corresponds to \cref{eq:bayes_sfs} with a uniform prior. \textbf{(3) Min:} $r_d(s) = \min_{z\in [1,..,n-1]}  -\psi^z \cdot \phi(s).$  \textbf{(4) Average:} $r_d(s) = -\frac{1}{n-1} \sum_{j=1}^{n-1}\psi^j \cdot \phi(s).$ \textbf{(5) None:} $r_d(s)=0$ or no diversity.

\begin{figure}[h]
\centering

\begin{subfigure}{0.49\textwidth}
  \begin{adjustbox}{width=.24\textwidth,valign=B}
    \begin{tabular}{lcr}
        \toprule
        \# & $r_e$ & \% \\
        \midrule
        \textbf{1} & 920 & 100 \\
        2 & 809 & 88 \\
        3 & 820 & 89 \\
        4 & 878 & 95 \\
        5 & 818 & 89 \\
        6 & 818 & 89 \\
        7 & 490 & 53 \\
        8 & 926 & 101 \\
        \bottomrule
  \end{tabular}
  \end{adjustbox}
        \includegraphics[width=.73\textwidth]{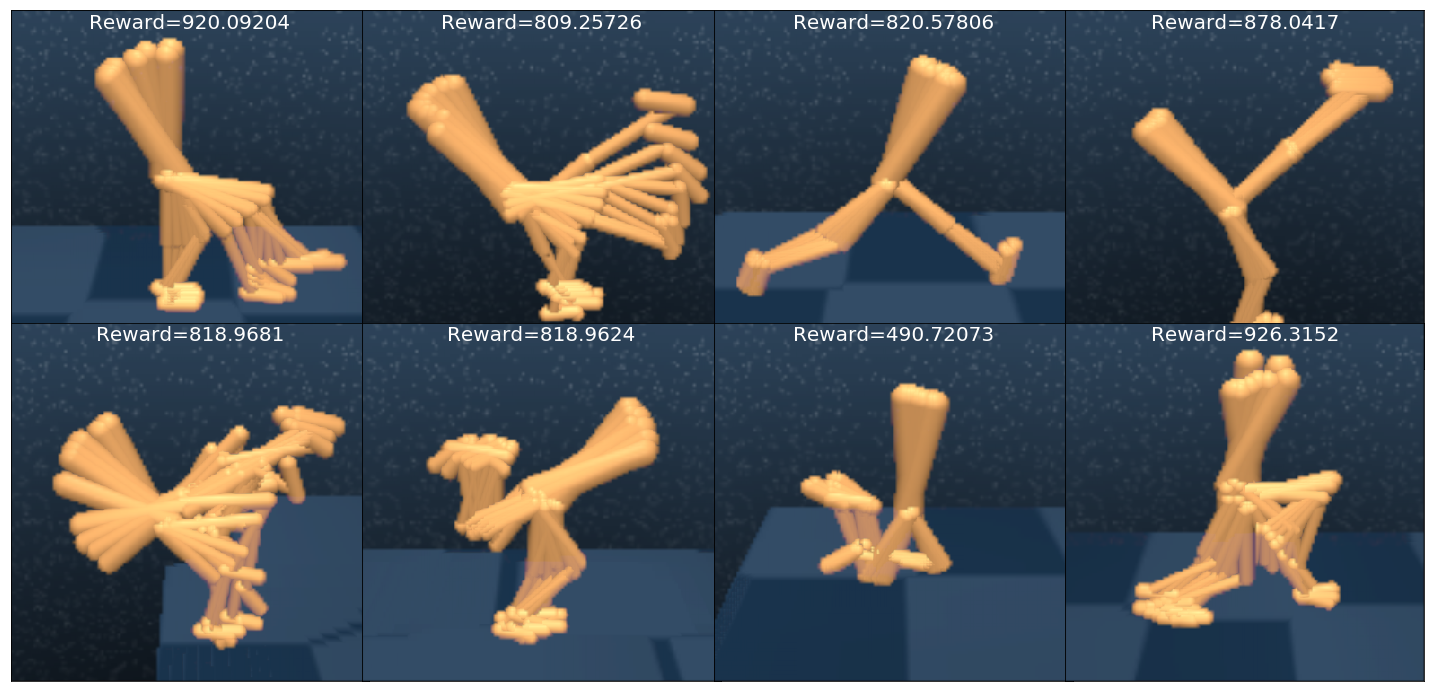}
         \caption{Walker Stand, $r_e$ as reward; $r_d$ as min.}
         \label{fig:walker_stand_min}
     \end{subfigure}
\begin{subfigure}{0.49\textwidth}
        \includegraphics[width=.73\textwidth]{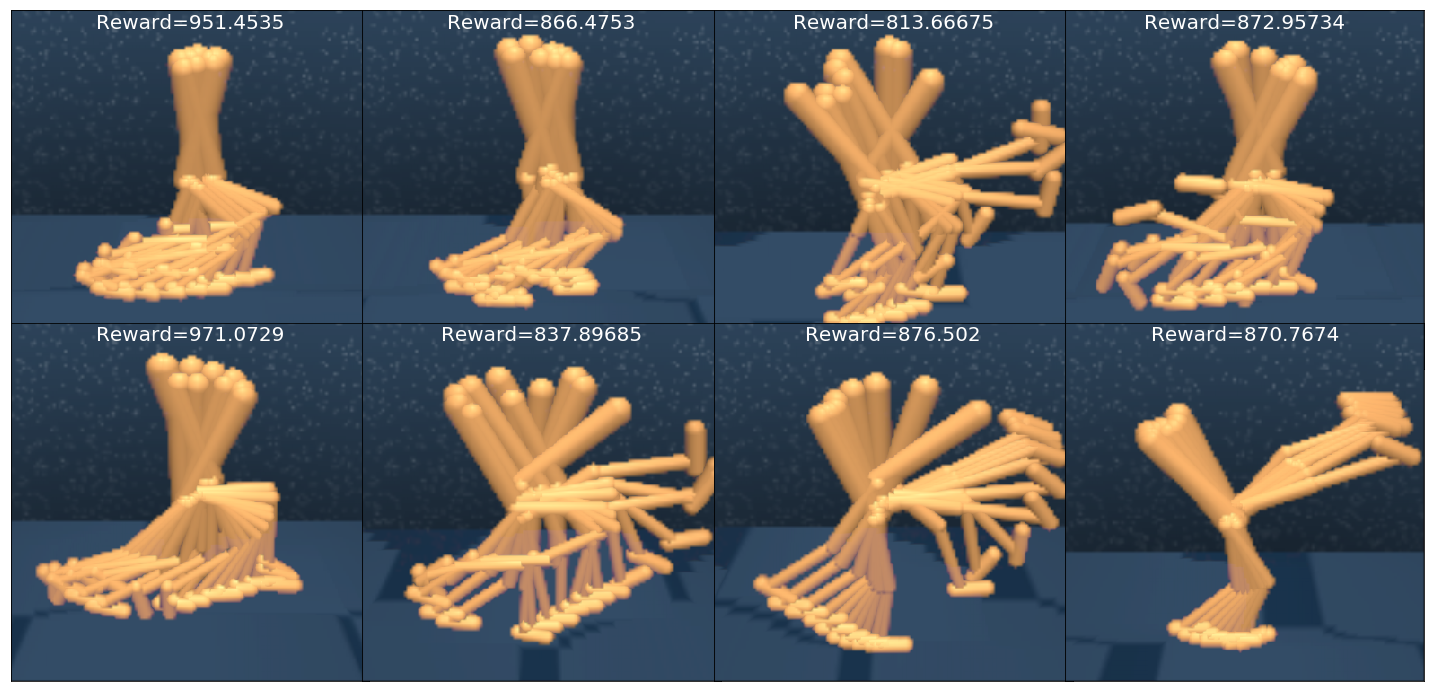}
          \begin{adjustbox}{width=.24\textwidth,valign=B}
    \begin{tabular}{lcr}
        \toprule
        \# & $r_e$ & \% \\
        \midrule
        \textbf{1} & 951 & 100 \\
        2 & 866 & 91 \\
        3 & 813 & 85 \\
        4 & 872 & 92 \\
        5 & 971 & 102 \\
        6 & 837 & 88 \\
        7 & 876 & 92 \\
        8 & 870 & 91 \\
        \bottomrule
  \end{tabular}
  \end{adjustbox}
         \caption{Walker Walk,  $r_e$ as reward; $r_d$ as average.}
         \label{fig:walker_walk_avg}
     \end{subfigure}
 \caption{Diverse near optimal policies in Walker}
\end{figure}

\cref{fig:walker_stand_min} presents eight polices that were discovered by \cref{alg:qd} where $r_d$ is the minimum explicit diversity criteria for \textbf{Walker.stand}. As we can see, the policies exhibit different types of standing: standing on both legs, standing on either leg, lifting the other leg forward and backward, spreading the legs and stamping. Not only are the policies different from each other, they also achieve high extrinsic reward in standing (see values on top of each policy visualization). Similar figures for the other diversity mechanisms can be found in the supplementary material (\cref{subsec:walker_stand}).
We observed that in this domain the Average diversity criterion can also discover policies that behave differently, but they are not as diverse as the ones found using the Minimum criterion (see \cref{subsec:walker_stand}  in the supplementary material)

The robustness mechanism can also provide diverse policies, but it tends to converge after a few iterations so no further diversity is achieved by the algorithm after 3 iterations. We also include a figure of different policies with no diversity mechanism in the supplementary (\cref{fig:supp_walker_stand_none}); in this case there is a small amount of diversity from training, but it is much less significant than the diversity we get with a diversity objective. Similarly, the discrimination method exhibits diversity but not as good as the explicit methods. We believe that this is due to the fact that the policies that maximize the extrinsic reward are already discriminative, and the algorithm fails to escape these local minima. 

\cref{fig:walker_walk_avg} presents similar results in the \textbf{Walker.walk} environment where $r_d$ is the average explicit diversity criteria. In this case the walker discovered how to walk in different ways, such as lifting one of the legs while up walking, walking with high knees, or walking with the heels to the bottom. In this domain we observed much better diversity with the explicit diversity mechanisms than with robustness or discrimination, see \cref{subsec:walker_walk}.  We also note that in both of the Walker environments, all (but one) of the discovered policies that we found are indeed near optimal, and satisfy the constraint (which was set to $90\%$). 


\begin{figure}[h]
\centering

\begin{subfigure}{0.49\textwidth}
  \begin{adjustbox}{width=.24\textwidth,valign=B}
    \begin{tabular}{lcr}
        \toprule
        \# & $r_e$ & \% \\
        \midrule
        \textbf{1} & 921 & 100 \\
        2 & 870 & 94 \\
        3 & 879 & 95 \\
        4 & 909 & 98 \\
        5 & 944 & 102 \\
        6 & 975 & 106 \\
        7 & 938 & 102 \\
        8 & 930 & 101 \\
        \bottomrule
  \end{tabular}
  \end{adjustbox}
        \includegraphics[width=.73\textwidth]{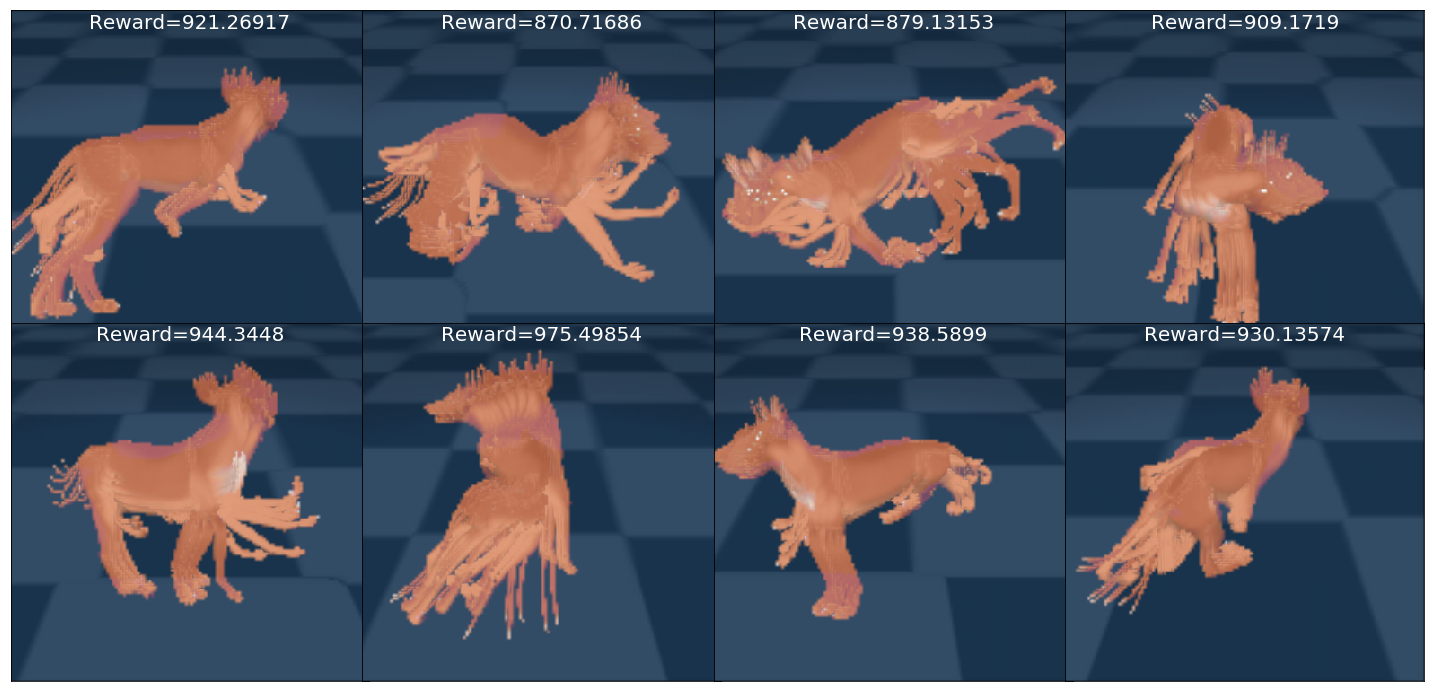}
         \caption{Dog Stand, $r_e$ as reward; $r_d$ as min.}
         \label{fig:dog_stand_min}
     \end{subfigure}
\begin{subfigure}{0.49\textwidth}
        \includegraphics[width=.73\textwidth]{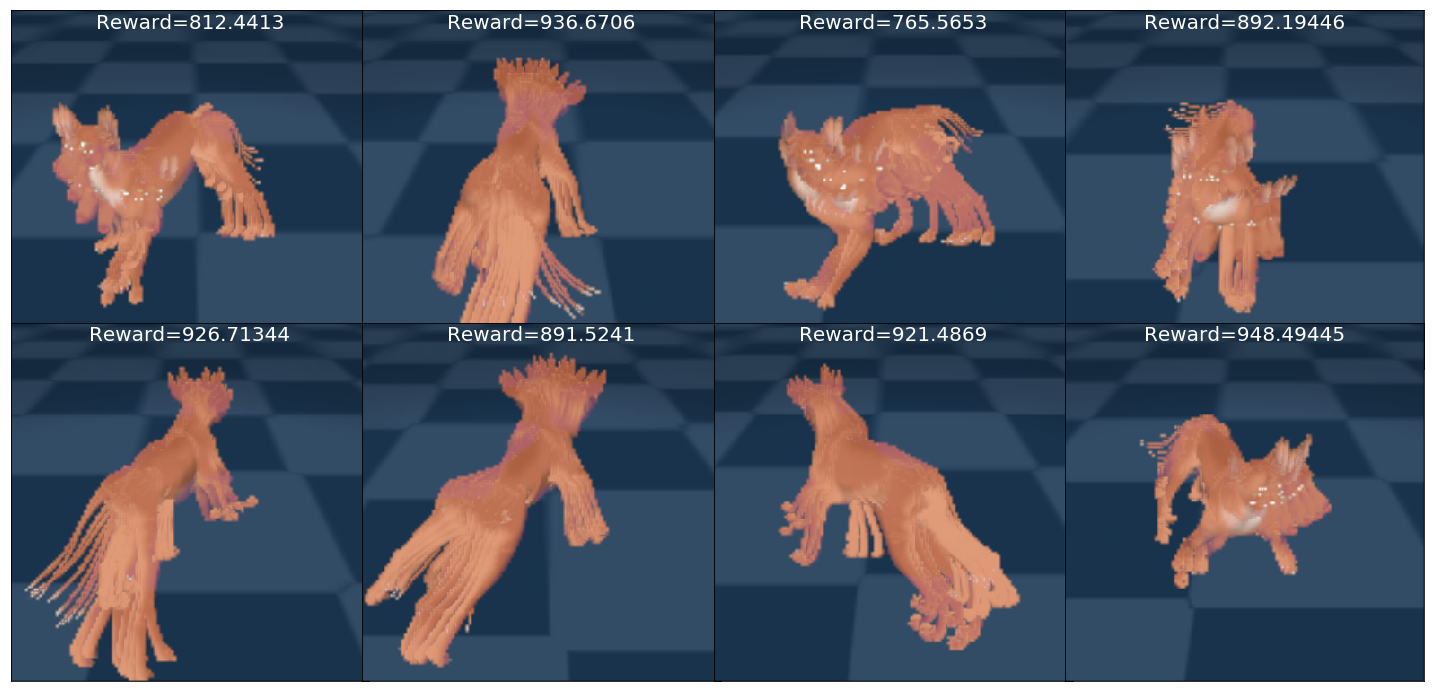}
          \begin{adjustbox}{width=.23\textwidth,valign=B}
    \begin{tabular}{lcr}
        \toprule
        \# & $r_e$ & \% \\
        \midrule
        \textbf{1} & 812 & --- \\
        2 & 936 & --- \\
        3 & 765 & --- \\
        4 & 892 & --- \\
        5 & 926 & --- \\
        6 & 891 & --- \\
        7 & 921 & --- \\
        8 & 948 & --- \\
        \bottomrule
  \end{tabular}
  \end{adjustbox}
         \caption{Dog Stand,  $r_e$ as reward; $r_d$ as none.}
         \label{fig:dog_stand_none}
     \end{subfigure}
 \caption{Diverse near optimal policies in Dog}
 \label{fig:dog}
\end{figure}

\cref{fig:dog} presents results in the \textbf{Dog.stand} environment where in \cref{fig:dog_stand_min} $r_d$ is the minimum explicit diversity criteria and in \cref{fig:dog_stand_none} there is no diversity mechanism.  Inspecting \cref{fig:dog_stand_none} we can see that the dog learns how to stand (different policies are independent of each other so we leave the $\%$ blank), but in all cases, it stands with four legs on the ground. On the other hand, in \cref{fig:dog_stand_min} the dog learns different variations of "three leg standing" (lifting one of his legs), and still achieves high reward.

Next, we present results in the \textbf{no-reward} setting, where the agent has no access to the reward from the environment. Our results with None diversity confirm that the implementation of these diversity mechanisms yields complex locomotion in the no-reward setting as was reported in the original papers. However, in more complex domains like Walker, without adding the explicit diversity we get static behaviours that resemble "Yoga" exercises, as was also reported, for example, in \citep{zahavy2021discovering}.

\begin{figure}[h]
\centering
\begin{subfigure}{0.49\textwidth}
\includegraphics[width=.9\textwidth]{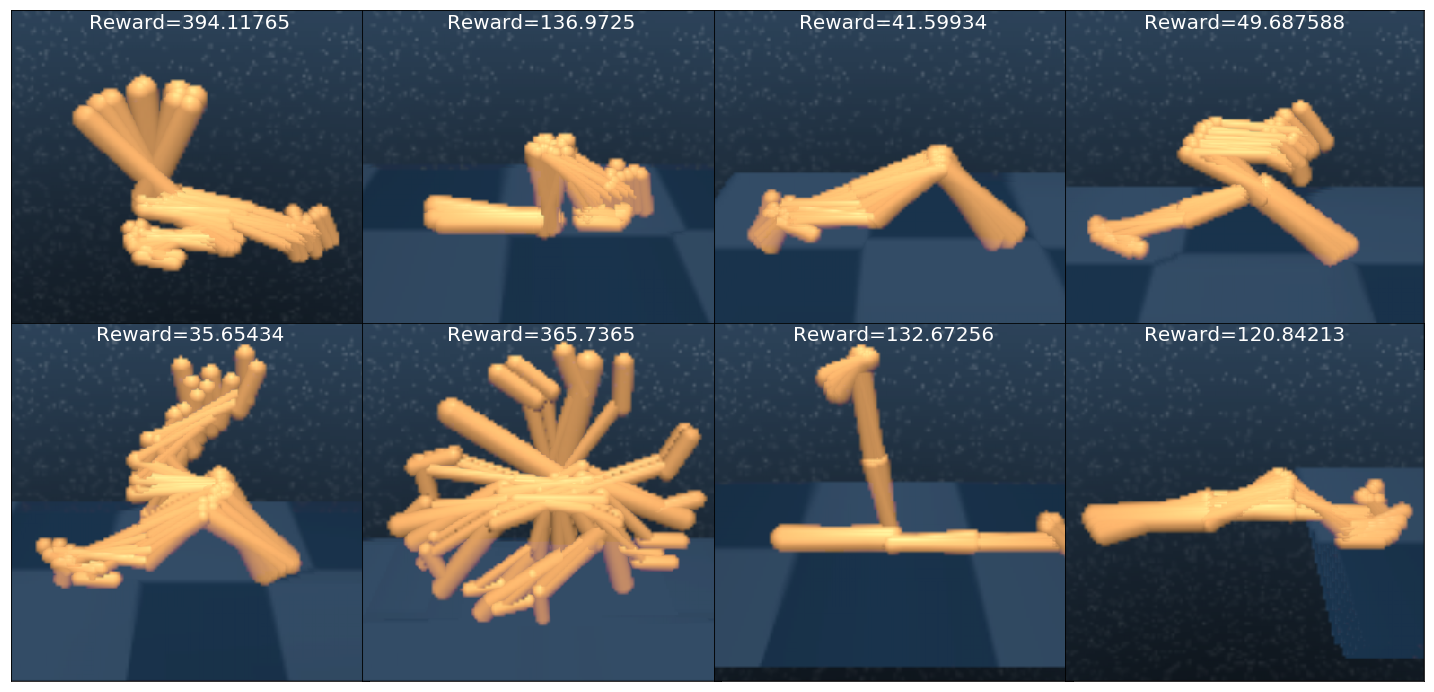}
\caption{Walker, $r_e$ as robustness and $r_d$ as average.}
\label{fig:walker_srand_smp_avg}
\end{subfigure}
\begin{subfigure}{0.49\textwidth}
\includegraphics[width=.9\textwidth]{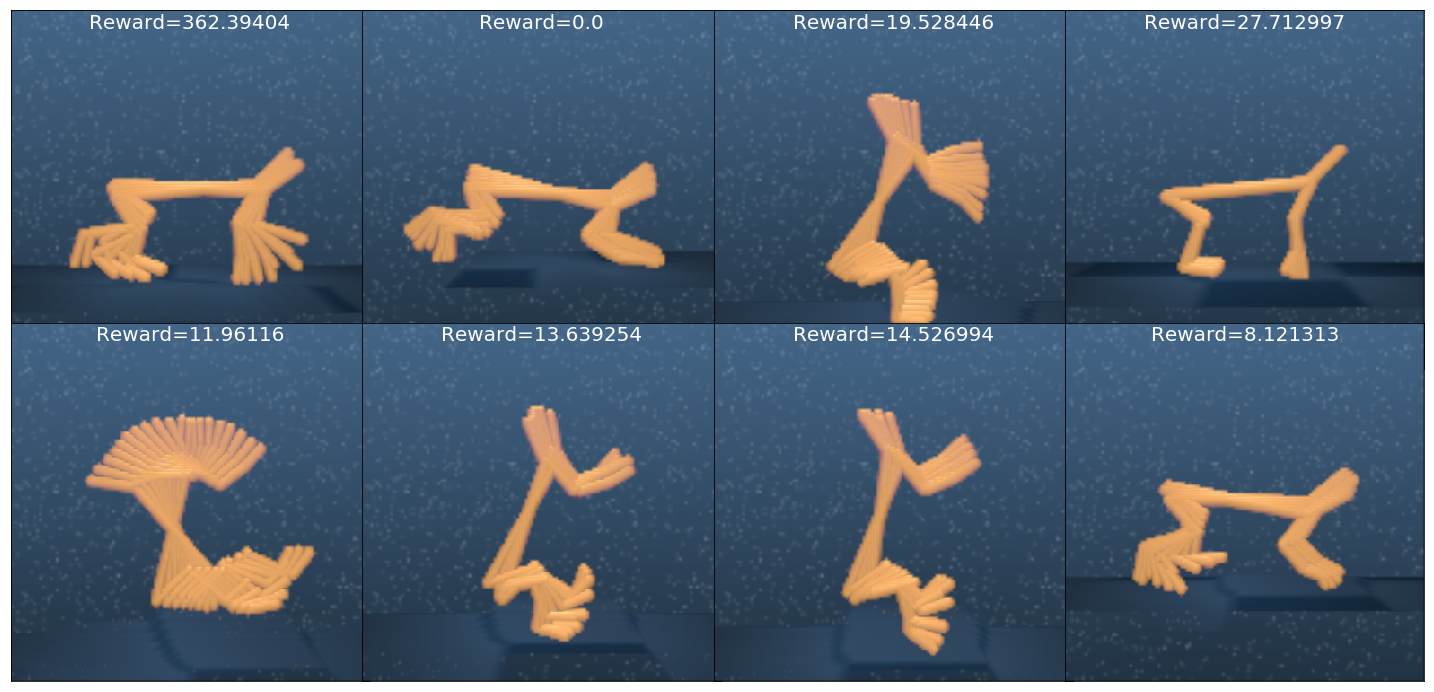}
\caption{Cheetah, $r_e$ as discrimination and $r_d$ as robustness.}
\label{fig:cheetah_run_vic_smp}
\end{subfigure}
\caption{Diversity without reward in Walker and Cheetah.}

\end{figure}

\cref{fig:walker_srand_smp_avg} presents results for \textbf{Walker} where $r_e$ is robustness and $r_d$ is average. Inspecting the results, we can see that the agent discovered complex locomotion skills such as kneeling backwards, crawling and flick-flack jumping. We also report the extrinsic reward for standing as another measure of zero-shot transfer (it was not used during training at all). In \cref{subsec:walker_smp} we can see that other diversity mechanisms discovered other surprising skills such as "head walking".

Finally, \cref{fig:cheetah_run_vic_smp} presents results for \textbf{Cheetah} where $r_e$ is discrimination and $r_d$ is robustness. The cheetah learns to run forward, backwards, and then to do various jumps. While previous methods were able to discover similar behaviours, they are typically not that diverse with such a small set.

\section{Conclusion}
In this work we proposed a framework for discovering near optimal diverse behaviours. We framed the problem as solving a CMDP where a diversity intrinsic reward and the extrinsic reward are adaptively combined. There are interesting connections to whitebox metagradients \citep{xu2018meta,zahavy2020self} -- the updates of the Lagrangian can be viewed as the outer update in metagradients where satisfying the constraint is the outer loss. Using metagradients to learn other diversity hyperparameters or even to discover the diversity reward itself \citep{zheng2018learning} are exciting directions for future work. Key to our approach was the idea of measuring diversity in the space of SFs. This design choice allowed us to provide insights on how existing diversity mechanisms behave from the perspective of convex optimization. 

There are many exciting applications for our framework. For example, consider the process of using RL to train a robot to walk. The designer does not know {\sl a priori} which reward will result in the desired walking pattern. Thus, robotic engineers often train a policy to maximize an initial reward, tweak the reward, and iterate until they reach the desired behaviour. Using our approach, the engineer would have multiple forms of walking to choose from in each attempt, which are also interpretable (linear in the weights). 


\clearpage

\bibliographystyle{abbrvnat}
\bibliography{ref}
     
\clearpage
\appendix

\section{Checklist}

\begin{enumerate}
\item For all authors

\begin{enumerate}
  \item Do the main claims made in the abstract and introduction accurately reflect the paper's contributions and scope?
    \answerYes
  \item Did you describe the limitations of your work?
    \answerYes{We discussed the limitations of diversity seeking methods, from being convex maximization problems (diayn) and from using auxiliary objectives (robustness). }
  \item Did you discuss any potential negative societal impacts of your work?
    \answerNo. Our paper studies how RL algorithms can find diverse solutions, we believe that promoting algorithmic diversity in AL should not have any negative societal impacts. 
  \item Have you read the ethics review guidelines and ensured that your paper conforms to them?
    \answerYes
\end{enumerate}

\item If you are including theoretical results...
\begin{enumerate}
  \item Did you state the full set of assumptions of all theoretical results?
    \answerYes
	\item Did you include complete proofs of all theoretical results?
    \answerYes
\end{enumerate}

\item If you ran experiments.
\begin{enumerate}
  \item Did you include the code, data, and instructions needed to reproduce the main experimental results (either in the supplemental material or as a URL)?
    \answerNo
  \item Did you specify all the training details (e.g., data splits, hyperparameters, how they were chosen)?
    \answerYes
	\item Did you report error bars (e.g., with respect to the random seed after running experiments multiple times)?
    \answerNo{Instead of repeating the same experiment over multiple seeds, each of our experiments was performed to discover eight policies sequentially. Our only numerical claim in this paper is about constraint satisfaction, and as our results suggest, it is being satisfied, in any of these eight, consecutive but independent trials (the parameters were initialized after each iteration. Another axis in which we tested our algorithm was the domain. So instead of repeating Walker.Walk a few times, we performed the second experiment on Walker.Stand (so, a diff in the extrinsic reward) and the same in Dog.}
	\item Did you include the total amount of compute and the type of resources used (e.g., type of GPUs, internal cluster, or cloud provider)?
    \answerNo
\end{enumerate}

\item If you are using existing assets (e.g., code, data, models) or curating/releasing new assets...
\begin{enumerate}
  \item If your work uses existing assets, did you cite the creators?
    \answerYes
  \item Did you mention the license of the assets?
    \answerNo
    \item Did you include any new assets either in the supplemental material or as a URL?
    \answerNo
  \item Did you discuss whether and how consent was obtained from people whose data you're using/curating?
    \answerNo
  \item Did you discuss whether the data you are using/curating contains personally identifiable information or offensive content?
    \answerNo
\end{enumerate}

\item If you used crowdsourcing or conducted research with human subjects...
\begin{enumerate}
  \item Did you include the full text of instructions given to participants and screenshots, if applicable?
    \answerNo
  \item Did you describe any potential participant risks, with links to Institutional Review Board (IRB) approvals, if applicable?
    \answerNo
    \item Did you include the estimated hourly wage paid to participants and the total amount spent on participant compensation?
    \answerNo
    \end{enumerate}

\end{enumerate}

\newpage
\section{Proof for \cref{lemma:conex_vic}}
\label{sec:proofvic}
\begin{proof}
We will focus on the case that there are no zero elements in $d_\pi$ which is a standard assumption in ergodic MDPs. Under this assumption $f$ is a twice differentiable function so it is convex if its Hessian is positive semidefinite. 

Recall that the prior $p(z)$ is constant, and that the policies $k = 1, ..., k\neq z$ are also constant from the perspective of $\dpiz$. We can therefore introduce a simplified notation and write the objective as 

$$
\sum_{s} \dpiz(s)  \log \left( \frac{\dpiz(s)p}{\dpiz(s)p + c_s}\right)
$$

The variable $\dpiz$ is a vector in the $|S|-$simplex. We can represent it using $|S|-1$ degrees of freedom $x_1, .. x_{s-1} \in [0,1]$ where the last element is $x_s = 1-\sum_{i=1}^{|S|-1} x_i.$ Notice that $x_s$ is a function of $x_i$ so it has a derivative with respect to. $x_i$ which equals $-1$. So we have
$$
f(x) = \sum_{i} x_i  \log \left( \frac{x_i p}{x_i p + c_i}\right)
+x_s\log \left( \frac{x_s p}{x_s p + c_s}\right)$$

The first derivative of this function with respect to $x_i, i\in[1,..,|S|-1$ is
\begin{align}
    \frac{\partial f}{\partial x_i}& = \log (x_i) + 1 + \log (p) - \frac{x_i p }{x_ip+ci} - \log (x_ip+c_i) \nonumber\\& - \log (x_s) -1 - \log(p) + \frac{x_s p }{x_sp+c_s} + \log (x_sp+c_s)
    \nonumber\\ &= \log (x_i) - \log (x_ip+c_i) - \frac{x_i p }{x_ip+ci} \label{eq:xi}\\& - \left( \log (x_s) - \log (x_sp+c_i)  - \frac{x_s p }{x_sp+c_s}\right)\label{eq:xs}
\end{align}
We can see that the terms in \cref{eq:xi} depend only on $x_i$ and the terms in \cref{eq:xs} depend only on $x_s$. In addition, we will soon see that the derivatives of $x_s$ will be equal for any $j \in 1,...,s-1.$ These two observations imply that the Hessian will have the form of $$H = D + m \barbelow{1},$$ where $D$ is a diagonal matrix with derivatives of \cref{eq:xi} with respect to $x_i$ as it elements, $\barbelow{1}$ is a matrix of all ones, and $m$ is the derivative of \cref{eq:xs} with respect to $x_j$ which we will show to be equal for all $j.$ Notice that $\forall x, $ we have that $x^T (D + m \barbelow{1}) x = \sum D_i x_i^2 + m (\sum x_i)^2.$ This implies that in order for the Hessian to be positive definite, we only need to show that the elements of $d$ and the scalar $m$ are positive. The derivative of \cref{eq:xi} with respect to $x_i$ is
\begin{align}
    &\frac{1}{x_i} -\frac{p}{px_i+c_i} - \frac{p(px_i+ci)-p^2x_i}{(px_i+c_i)^2}  \nonumber\\& = 
    \frac{px_i+c_i - px_i}{x_i(px_i+c_i)}  - \frac{pc_i}{(px_i+c_i)^2}  \nonumber\\& = \frac{c_i(px_i+c_i)-px_ic_i}{x_i(px_i+c_i)^2} = \frac{c_i^2}{x_i(px_i+c_i)^2},
\end{align}
which is positive because $x_i \ge 0$. 

Similarly, The derivative of \cref{eq:xs} with respect to $x_j$ is
\begin{align}
    &\frac{1}{x_s} -\frac{p}{px_s+c_s} - \frac{p(px_s+cs)-p^2x_s}{(px_s+c_s)^2}  \nonumber\\& = 
    \frac{c_s^2}{x_s(px_s+c_s)^2},
\end{align}
which is also positive because $x_s \ge 0$ and concludes our proof. 
\end{proof}








\section{Proof for \cref{FW:wcpi} }
\label{sec:wcpi}

\begin{algorithm}[H]
\caption{The iterative procedure in \cite{zahavy2021discovering}}\label{alg:greedy-iterative}
\begin{algorithmic}
\STATE \textbf{Initialize:} Sample $\w \sim N(\bar 0,\bar 1), \Pi^0 \leftarrow \{\ \}, \pi^1 \leftarrow \arg\max_{\pi \in \Pi} \w \cdot \vpsi^\pi$, $t \leftarrow 1$ \\
\STATE $\bar{v}^{\text{SMP}}_{\Pi^1} \leftarrow  {-||\vpsi^1||}$
\REPEAT
    \STATE $\Pi^t \leftarrow \Pi^{t-1} \cup \{\pi^t\}$
    \STATE $\Psi^t = \Psi^{t-1} \cup \{\psi^t\}$
    \STATE $\bar{\w}^{\text{SMP}}_{\Pi^t}$ $\leftarrow \arg \min_{\w \in \mathbb{B}_2}   \max_{\psi \in \Psi^i} w \cdot \psi$\\
    \STATE $\pi^{t+1}$ $\leftarrow\arg\max_{\pi} \psi(\pi) \cdot \bar{\w}^{\text{SMP}}_{\Pi^t}$ \\
    \STATE $t \leftarrow t+1$
\UNTIL{${v}^{t}_{\wapi} \le \bar{v}^{\text{SMP}}_{\Pi^{t-1}}$}
\STATE \textbf{return} $\Pi^{t-1}$
\end{algorithmic}
\end{algorithm}

\begin{algorithm}[H]
\caption{Fully corrective FW for $h(\psi) = 0.5||\psi||^2_2$}\label{alg:fw}
\begin{algorithmic}
\STATE \textbf{Initialize:} Let $\pi^1$ be a random policy and let $\psi^1$ be its SFs. Also, let $\Pi^0=\{\}$ and $\Psi^0 = \{\}$ and $t\leftarrow 1.$ \\
\REPEAT
    \STATE $\Pi^t = \Pi^{t-1} \cup \{\pi^t\}$
    \STATE $\Psi^t = \Psi^{t-1} \cup \{\psi^t\}$
    \STATE $\hat \psi = \arg\min_{\psi \in \text{Co} (\Psi^{i})} 0.5||\psi||^2_2.$
    \STATE $ \pi^{t+1} = \arg\max_{\pi} \psi(\pi) \cdot -\nabla h(\hat \psi) = \arg\max_{\pi} \psi(\pi) \cdot -\hat \psi $ \\
    \STATE $t \leftarrow t+1$
\UNTIL{$h(\hat\psi) \le \epsilon$}
\STATE \textbf{return} $\Pi^{t-1}$
\end{algorithmic}
\end{algorithm}

In this section we show that the iterates of the fully corrective FW algorithm (\cref{alg:fw}) correspond to the iterates of the Worst Case Policy Iteration algorithm (\cref{alg:greedy-iterative}). Examining the two algorithms, it is easy to see that all that is needed is to show that $$\arg\max_{\pi} \psi(\pi) \cdot -\hat \psi = \arg\max_{\pi} \psi(\pi) \cdot \wapi.$$

To show this, first observe that $\wapi$ can be also written as 
\begin{equation}
\label{eq:minmax}
    \wapi = \arg \min_{\w \in \mathbb{B}_2}   \max_{x\in \Psi^i} w \cdot \psi = \arg \min_{\w \in \mathbb{B}_2}   \max_{\psi \in \text{Co}(\Psi^i)} w \cdot \psi,
\end{equation}
that is, maximizing $\psi$ over $\text{Co}(\Psi^i)$ instead of $\Psi^i$ (SMP). This is correct because for any reward $w$ there is always a maximizer in the convex hull that is one of the vertices (a property of the linear inner product). And therefore, the same maximum value is attained when maximizing over these two sets.

Next, we have that 
\begin{align}
    \arg\min_{\psi\in \text{Co} (\Psi^{i})} ||\psi||^2_2 &=  \arg\max_{\psi\in \text{Co} (\Psi^{i})} -||\psi||^2_2 \\
    & =  \arg\max_{\psi\in \text{Co} (\Psi^{i})} -||\psi||_2 = \arg\max_{\psi\in \text{Co} (\Psi^{i})} \min_{w\in \mathcal{B}_2} \psi\cdot w. \label{eq:maxmin}
\end{align}

Now, if we denote the optimal solutions to \cref{eq:maxmin} as $\hat w, \hat \psi$ then, they are also an optimal solution to \cref{eq:minmax} via Von Neuman's min-max theorem. This means that $\wapi = \hat w = -\hat \psi / ||\hat \psi||.$ 

Thus $$
\arg\max_{\pi} \psi(\pi) \cdot \wapi = \arg\max_{\pi} \psi(\pi) \cdot -\hat \psi / ||\hat \psi|| = \arg\max_{\pi} \psi(\pi) \cdot -\hat \psi,
$$
where the second inequality follows from the fact that dividing the reward by the same constant across all states does not change the optimal policy (the $\arg\max$).

Finally, note that the function $h=0.5 ||x||^2_2$ has 1-Lipschitz gradient and is strongly convex. Thus, since the algorithms are equivalent, \cref{alg:greedy-iterative} achieves a linear convergence according to the following theorem. 

\begin{theorem}[Linear Convergence \citep{jaggi2015global}]
\label{thm:fw_lin}
Suppose that $h$ has L-Lipschitz gradient  and is $\mu$-strongly convex. Let $D = \{d_\pi, \forall \pi \in \Pi \}$ be the set of all the state occupancy's of deterministic policies in the MDP and let $\mathcal{K} = Co(D)$ be its Convex Hull. Such that $\mathcal{K}$ a polytope with vertices $D$, and let $M= diam(\mathcal{K})$. Also, denote the Pyramidal Width of $D,$  $\delta = PWidth(D)$ as in \citep[Equation 9 1]{jaggi2015global}.

Then the suboptimality $h_t$ of the iterates of all the fully corrective FW algorithm decreases geometrically at each step, that is
$$
h(x_{t+1}) \le (1 - \rho) h(x_{t}) \text{ , where } \rho = \frac{\mu \delta^2}{4L M^2}
$$
\end{theorem}

\section{Additional implementation details and hyper parameters}
\label{sec:tech_details}

When we add a new policy, $\pi^t,$ to the set $\Pi^{t-1}$, we reset the maximum value $\vem = \max \{\vem, v^t \}$. This step is useful because the policies and their value functions are computed approximately in practice and in some of the domains the optimal performance is not achieved in the first iteration of \cref{alg:qd}.

To bound the intrinsic rewards we first use the following transformation $\tilde r_w(s) = \frac{w\cdot\phi(s)\ + \|w\|^2}{\|w\|^2}$ and then apply the following non-linear transformation:   
\begin{equation}
    \label{eq:explicit_normed}
    r(s) = \left(1-\exp\left(-\tau \tilde r_w(s) \right)\right)/ (1-\exp(\tau)), 
\end{equation}
This transformation is useful when we want the reward to be more sensitive to small variations of the inner product, i.e., when many policies are relatively similar to each other.

Finally, \cref{table:hyperparameters} summarizes the hyperparameters that we use in \cref{alg:qd}
\begin{table}[h]
\caption{Hyperparameters table}
\begin{center}
\begin{tabular}{|l|l|}
    \hline
    Parameter & Value \\
    \hline 
    Optimality level $\alpha$ (\cref{eq:cmdp_L}) & 0.9 \\
    Environment steps per policy & $10^6$  \\
    Number of policies & $8$  \\
    Lagrange entropy regularization weight $a_h$ (\cref{eq:lagrange}) & $0.01$  \\
    Lagrange learning rate & $0.1$  \\
    Lagrange update frequency ($N_{\lambda}$) & $30$  \\
    Estimation decay factor $a_d$  & $0.9$  \\
    Normalization temperature $\tau$ (\cref{eq:explicit_normed}) & 3\\ 
    \hline
\end{tabular}
\end{center}
\label{table:hyperparameters}
\end{table}

\section{Additional results}

Our "motion figures" were created in the following manner. Given a trajectory of frames that composes a video $f_1,\ldots,f_T$, we first trim and sub sample the trajectory into a point of interest in time: $f_n,\ldots,f_{n+m}$. We always use the same trimming across the same set of policies (the sub figures in a figure). We then sub sample frames from the trimmed sequence at frequency $1/p$: $f_n, f_{n+p}, f_{n+2p}\ldots,$. After that, we take the maximum over the sequence and present this "max" image. In Python, this simply corresponds to, for example, to
\begin{align*}
    &\text{n=400, m=30, p=3}\\
    &\text{indices = range(n,n+m,p)}\\
    &\text{im} = \text{np.max(f[indices])}\\
\end{align*}
This creates the effect of motion in single figure since the object has higher values then the background.

\subsection{Clockwise Diversity in  Cartpole and Reacher}
\begin{figure}[h]
\centering
\begin{subfigure}{0.49\textwidth}
\includegraphics[width=\linewidth]{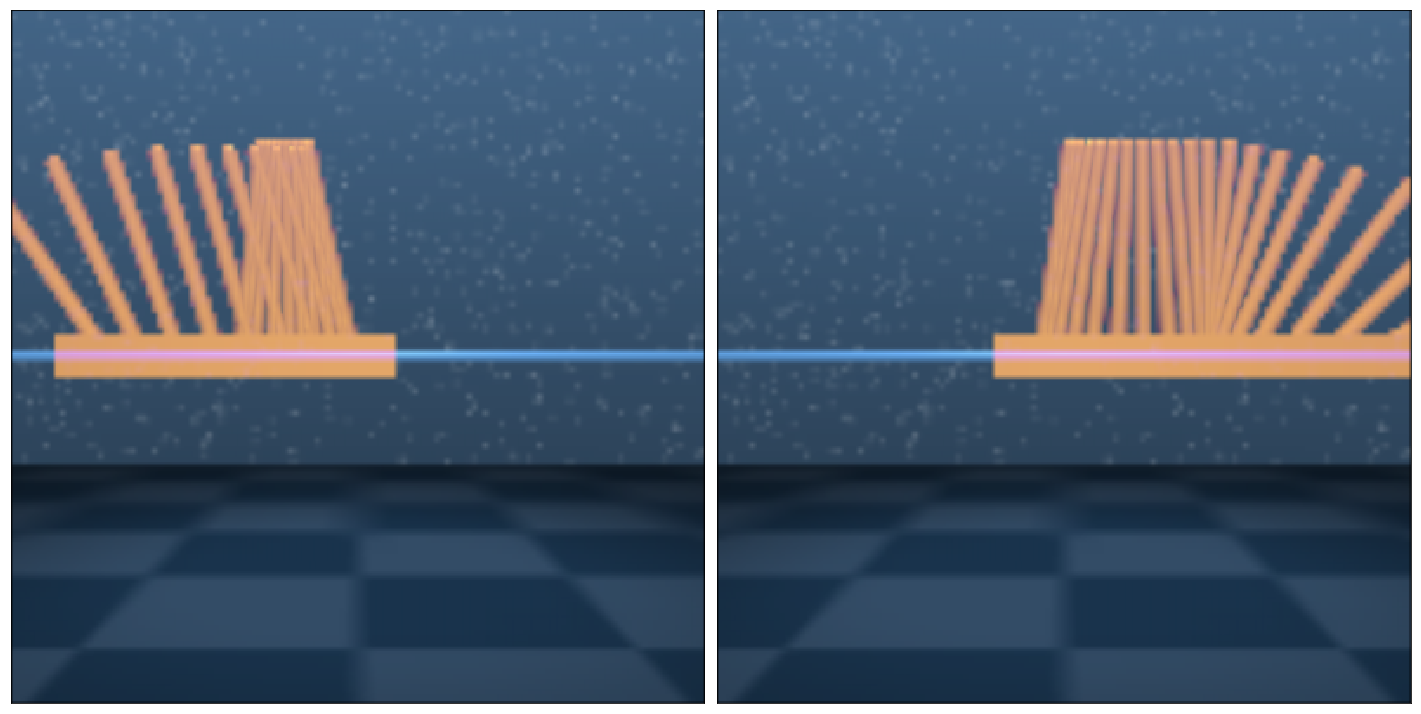}
\caption{Cartpole}
\end{subfigure}
\begin{subfigure}{0.49\textwidth}
\includegraphics[width=\linewidth]{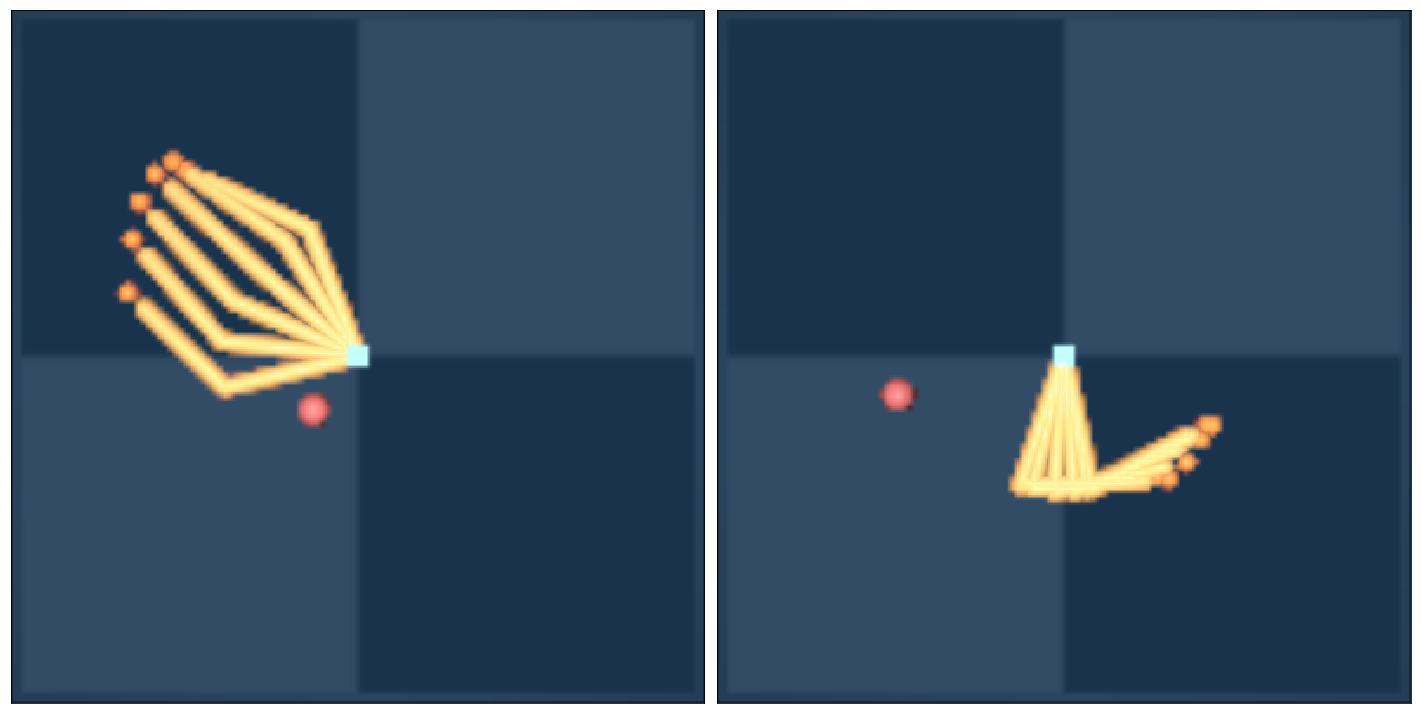}
\caption{Reacher}
\end{subfigure}
\caption{Clockwise Diversity in  Cartpole and Reacher. \label{fig:cartpole}}
\end{figure}

\clearpage
\subsection{Walker Stand}
\label{subsec:walker_stand}
\begin{figure}[h]
\centering
\includegraphics[width=0.9\linewidth]{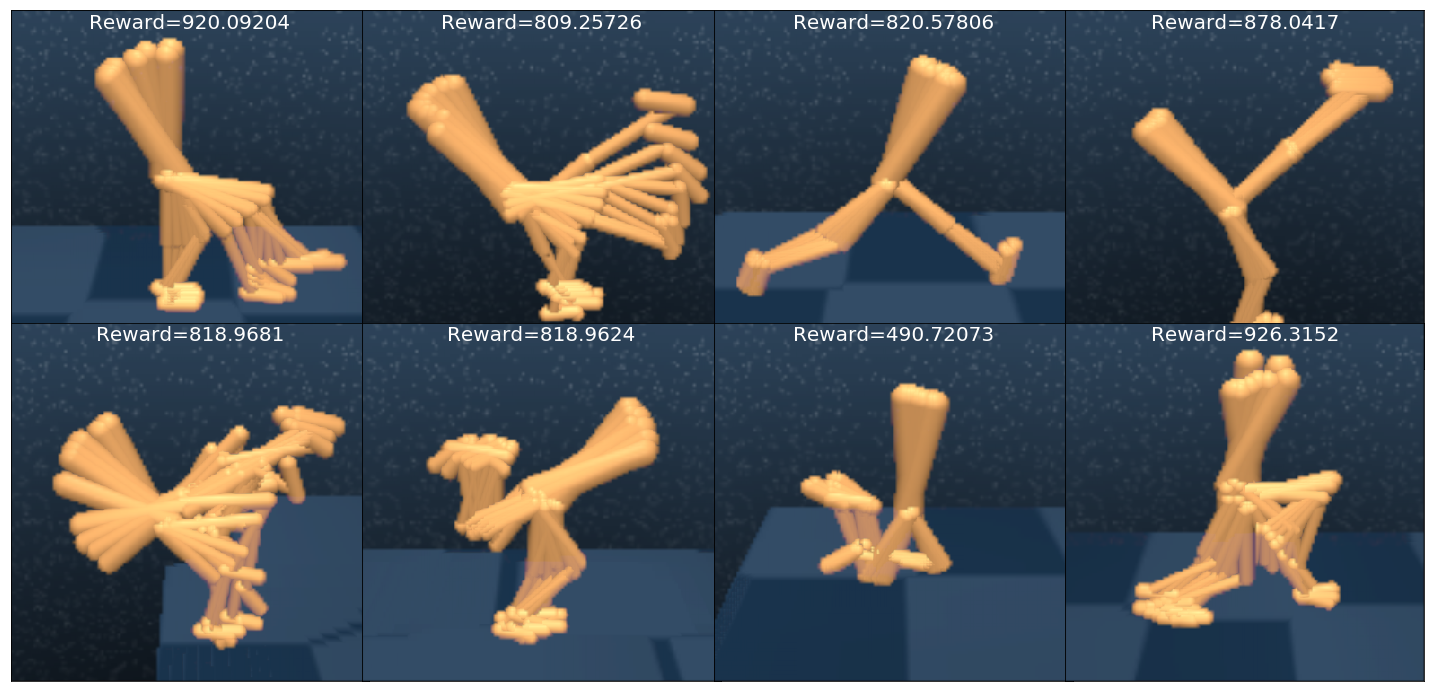}
\caption{Min}
\label{fig:supp_walker_stand_min}
\end{figure}
\begin{figure}[h]
\centering
\includegraphics[width=0.9\linewidth]{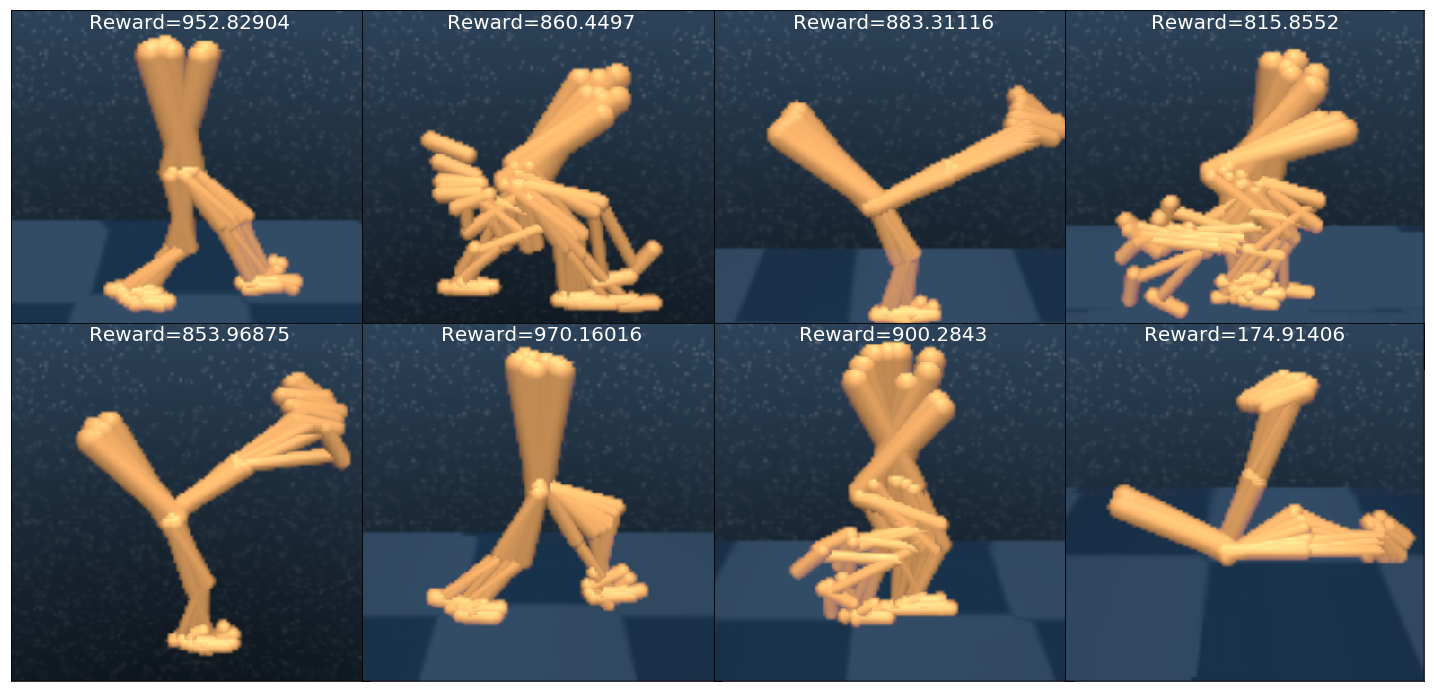}
\caption{Average}
\label{fig:supp_walker_stand_avg}
\end{figure}
\begin{figure}[h]
\centering
\includegraphics[width=0.9\linewidth]{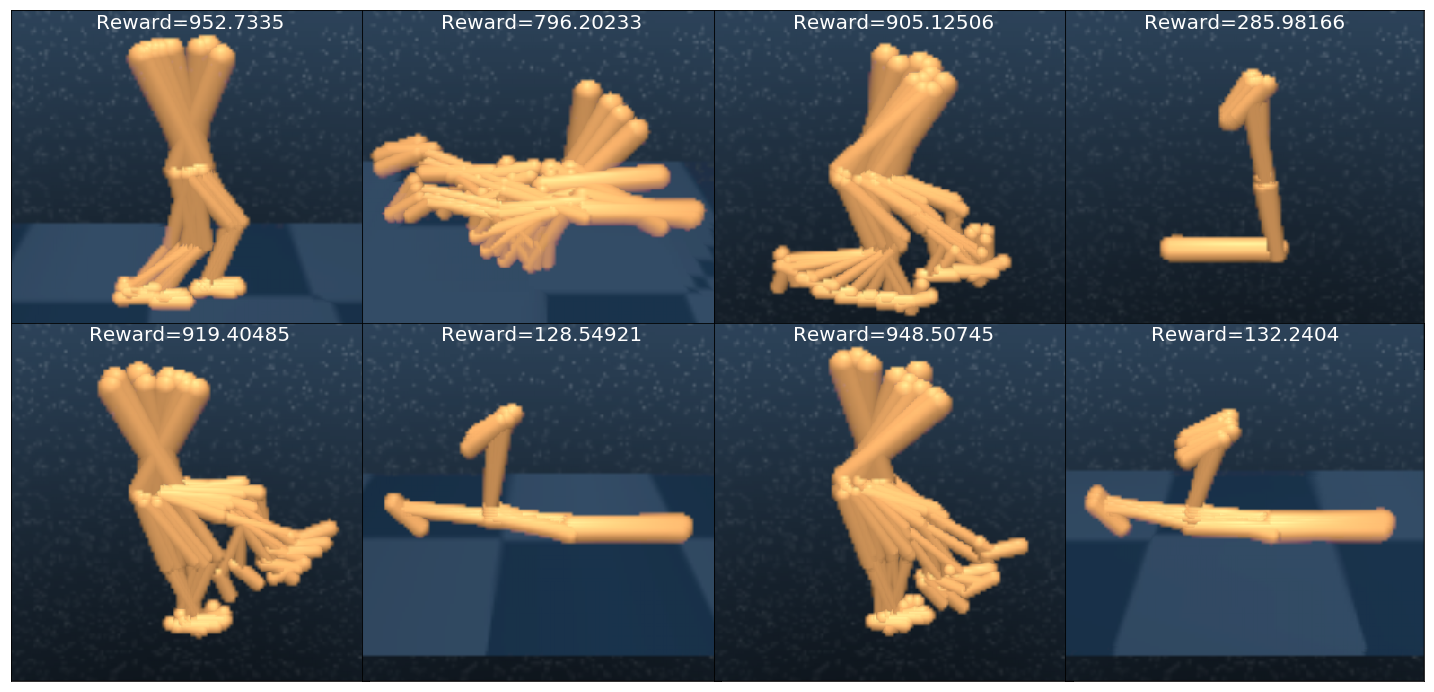}
\caption{Robustness}
\label{fig:supp_walker_stand_rob}
\end{figure}
\begin{figure}[h]
\centering
\includegraphics[width=0.9\linewidth]{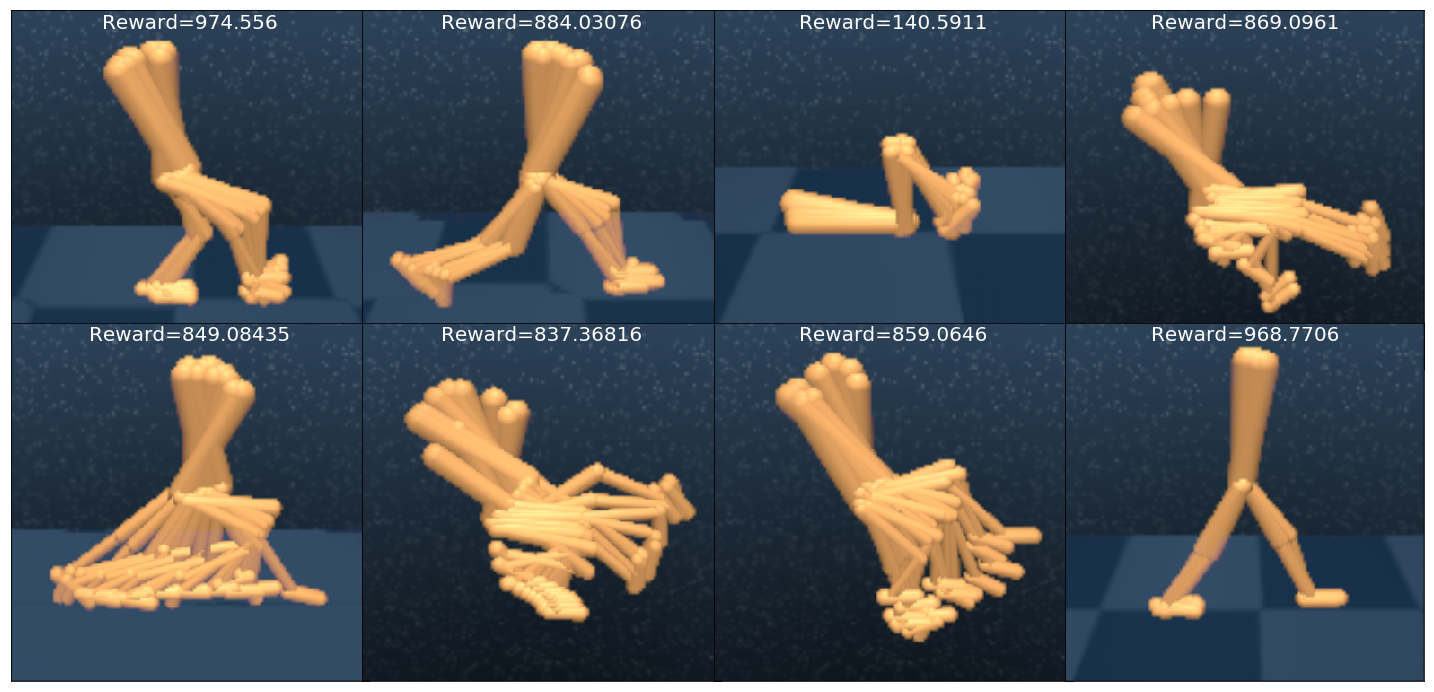}
\caption{Discrimination}
\label{fig:supp_walker_stand_disc}
\end{figure}
\begin{figure}[h]
\centering
\includegraphics[width=0.9\linewidth]{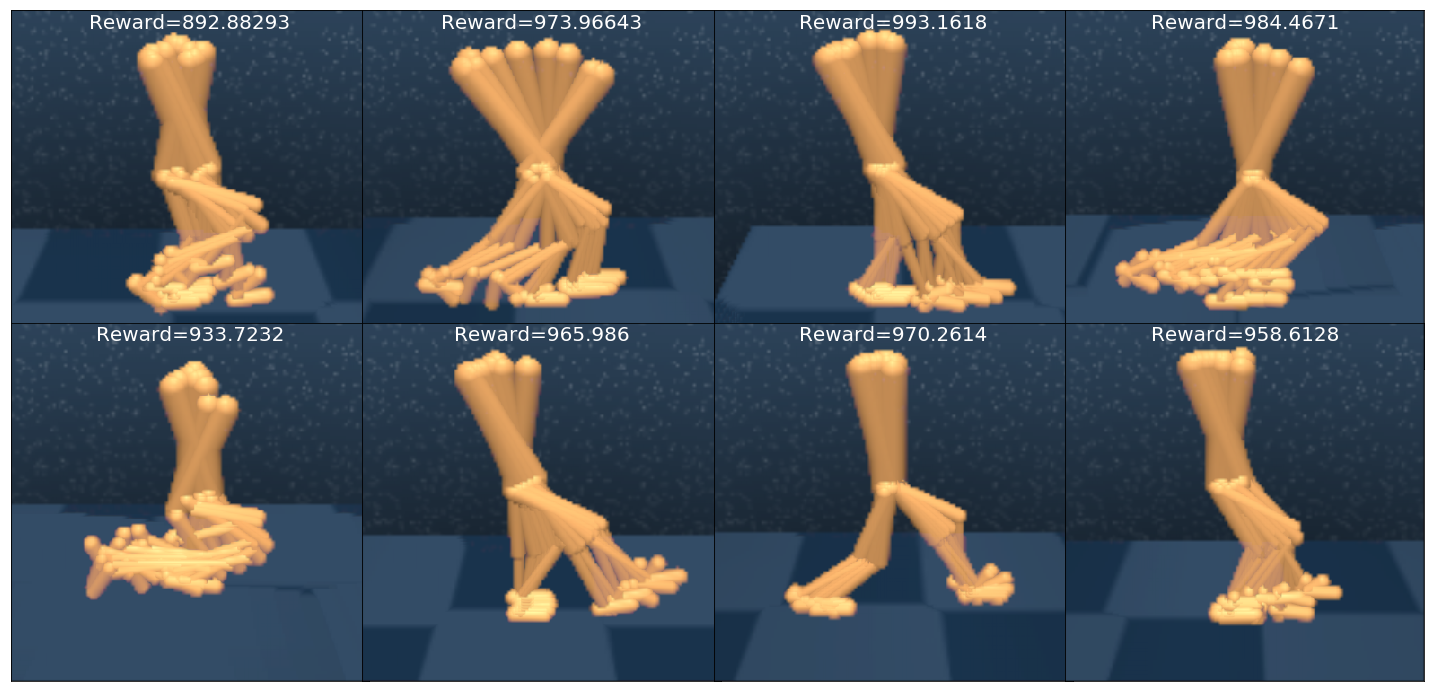}
\caption{None}
\label{fig:supp_walker_stand_none}
\end{figure}

\clearpage
\subsection{Walker Walk}
\label{subsec:walker_walk}
\begin{figure}[h]
\centering
\includegraphics[width=0.9\linewidth]{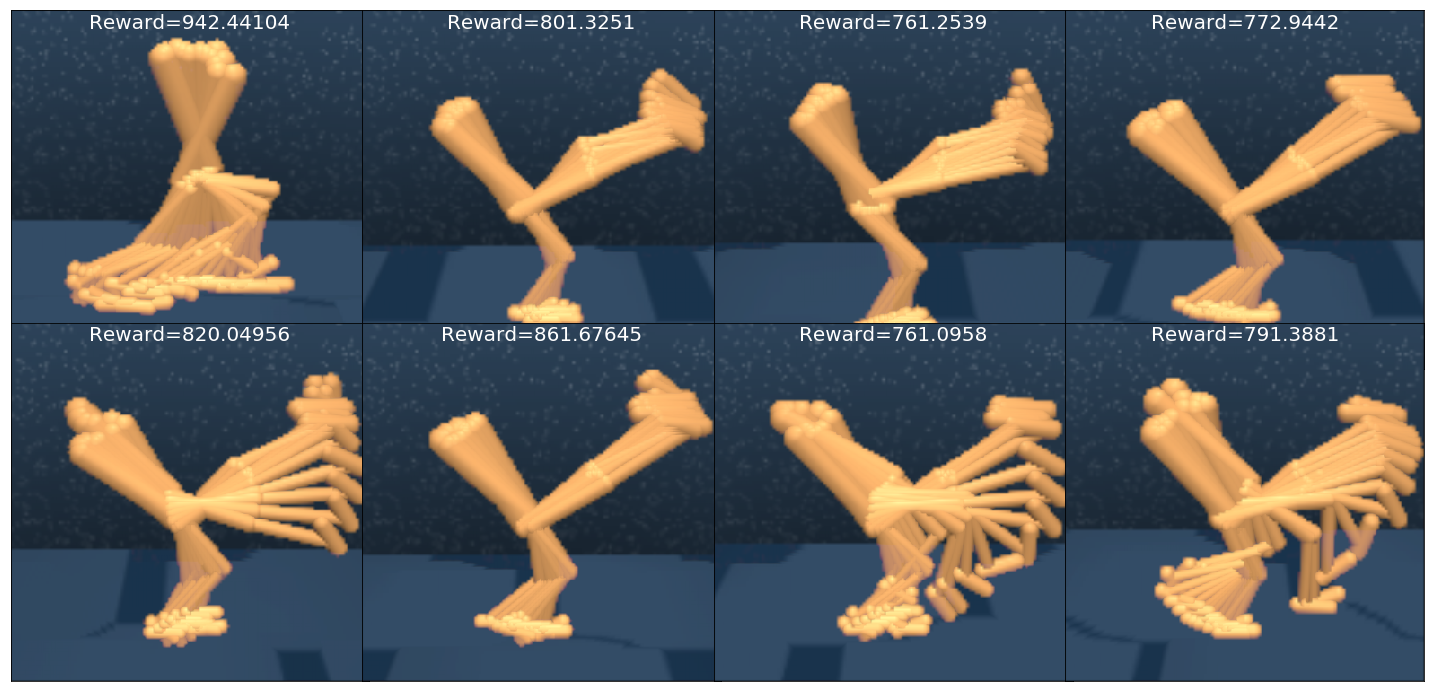}
\caption{Min}
\label{fig:supp_walker_walk_min}
\end{figure}
\begin{figure}[h]
\centering
\includegraphics[width=0.9\linewidth]{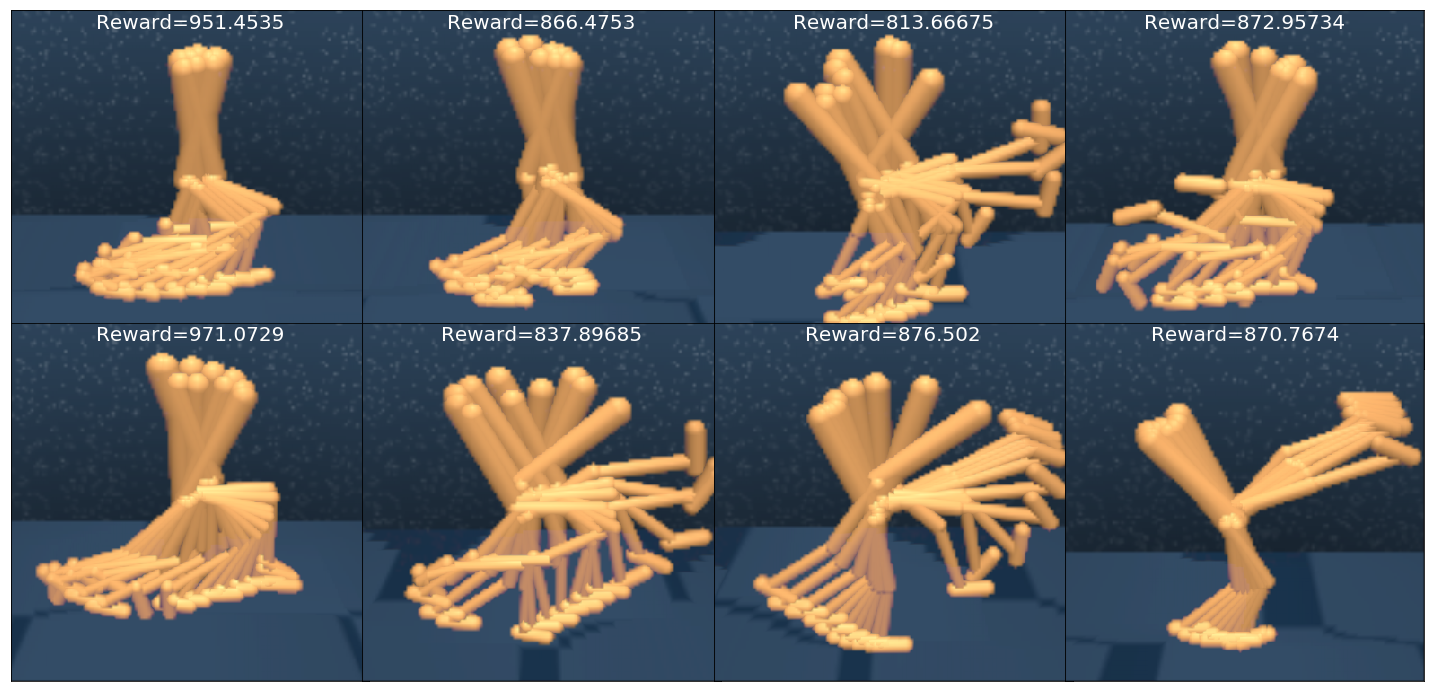}
\caption{Average}
\label{fig:supp_walker_walk_avg}
\end{figure}
\begin{figure}[h]
\centering
\includegraphics[width=0.9\linewidth]{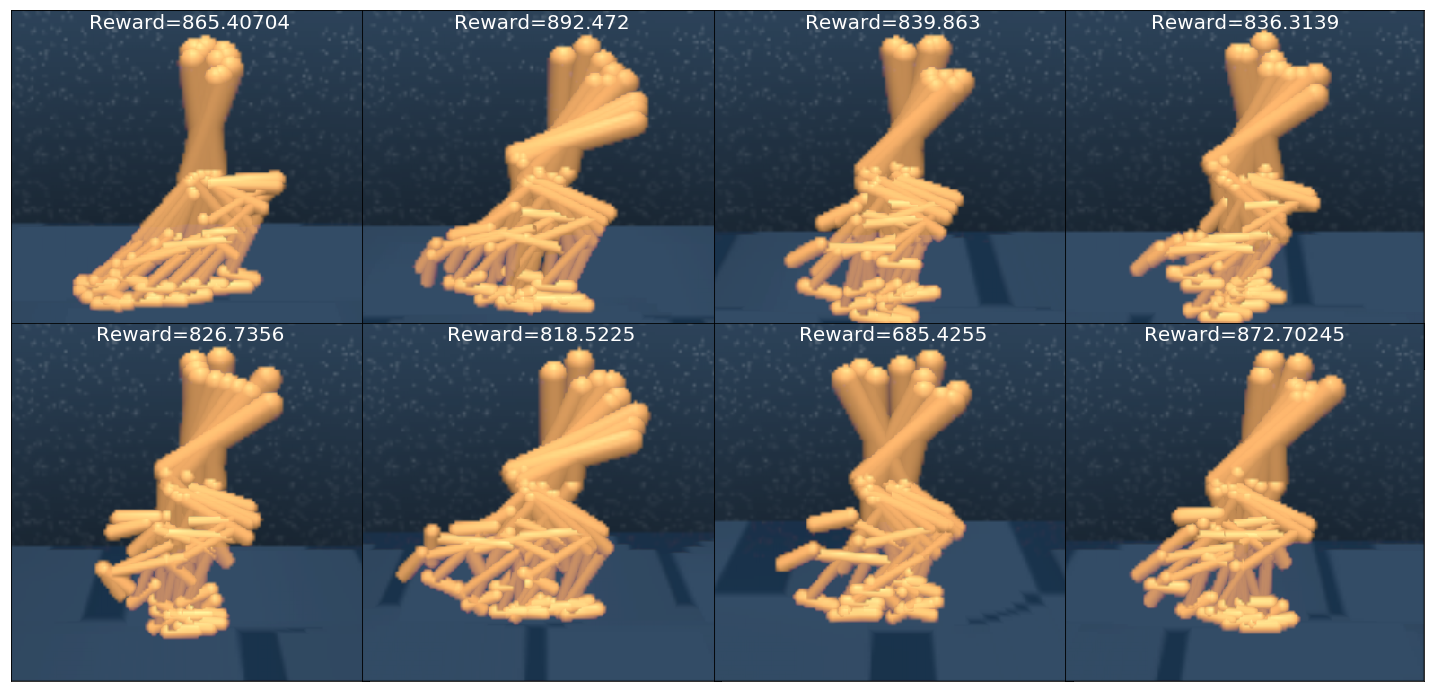}
\caption{Robustness}
\label{fig:supp_walker_walk_rob}
\end{figure}
\begin{figure}[h]
\centering
\includegraphics[width=0.9\linewidth]{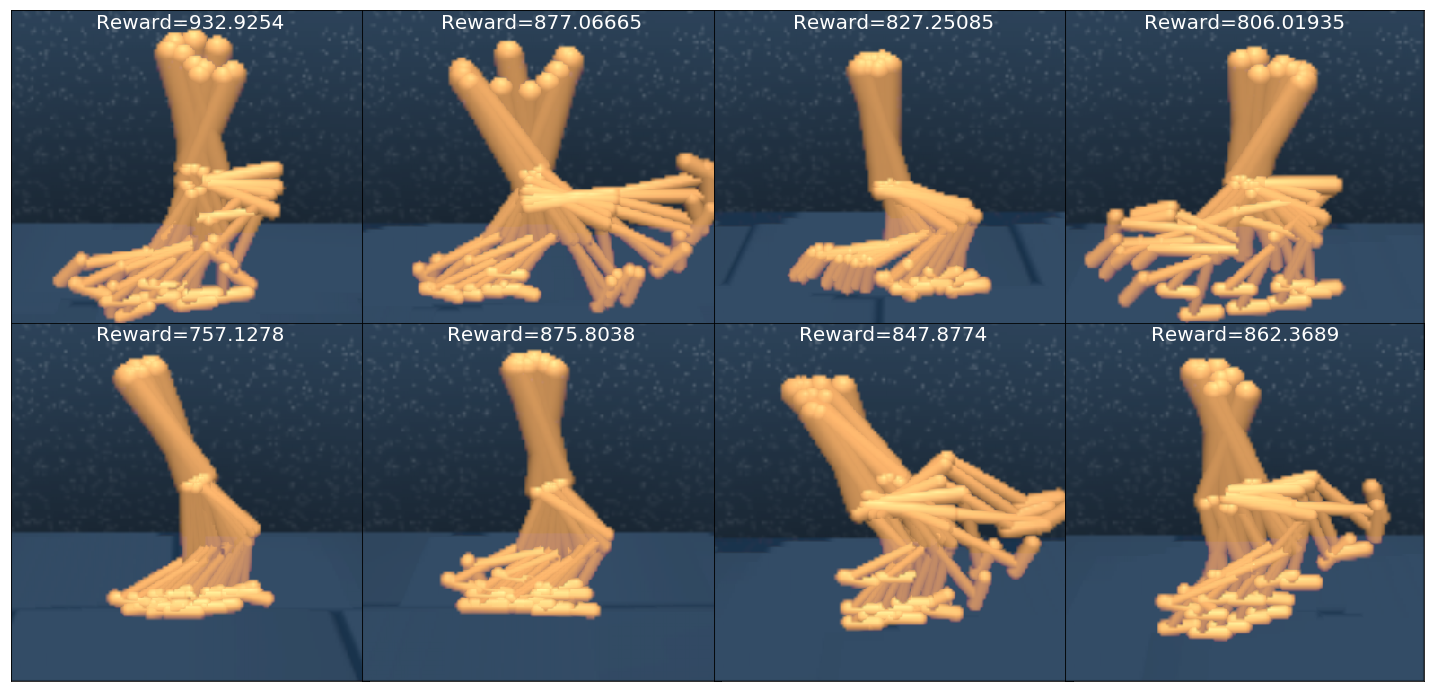}
\caption{Discrimination}
\label{fig:supp_walker_walk_disc}
\end{figure}
\begin{figure}[h]
\centering
\includegraphics[width=0.9\linewidth]{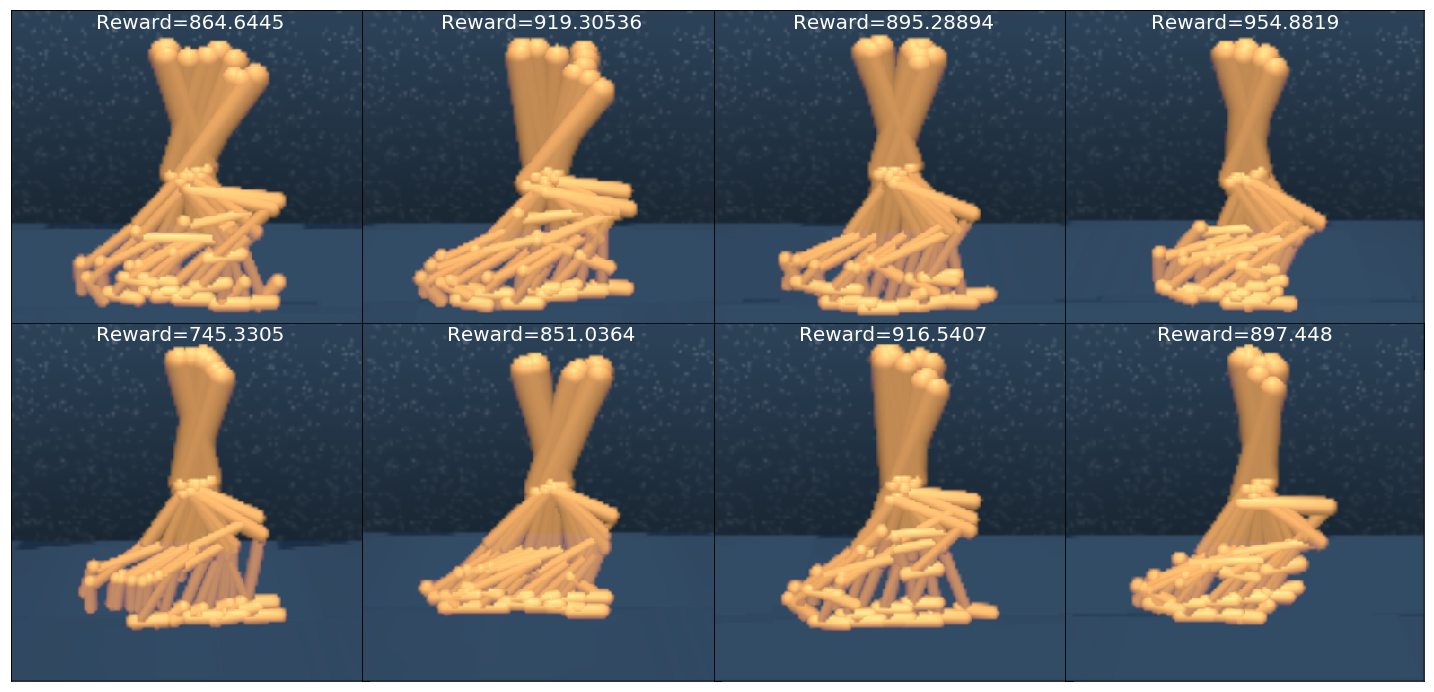}
\caption{None}
\label{fig:supp_walker_walk_none}
\end{figure}

\clearpage
\subsection{Robustness in Walker}
\label{subsec:walker_smp}
\begin{figure}[h]
\centering
\includegraphics[width=0.9\linewidth]{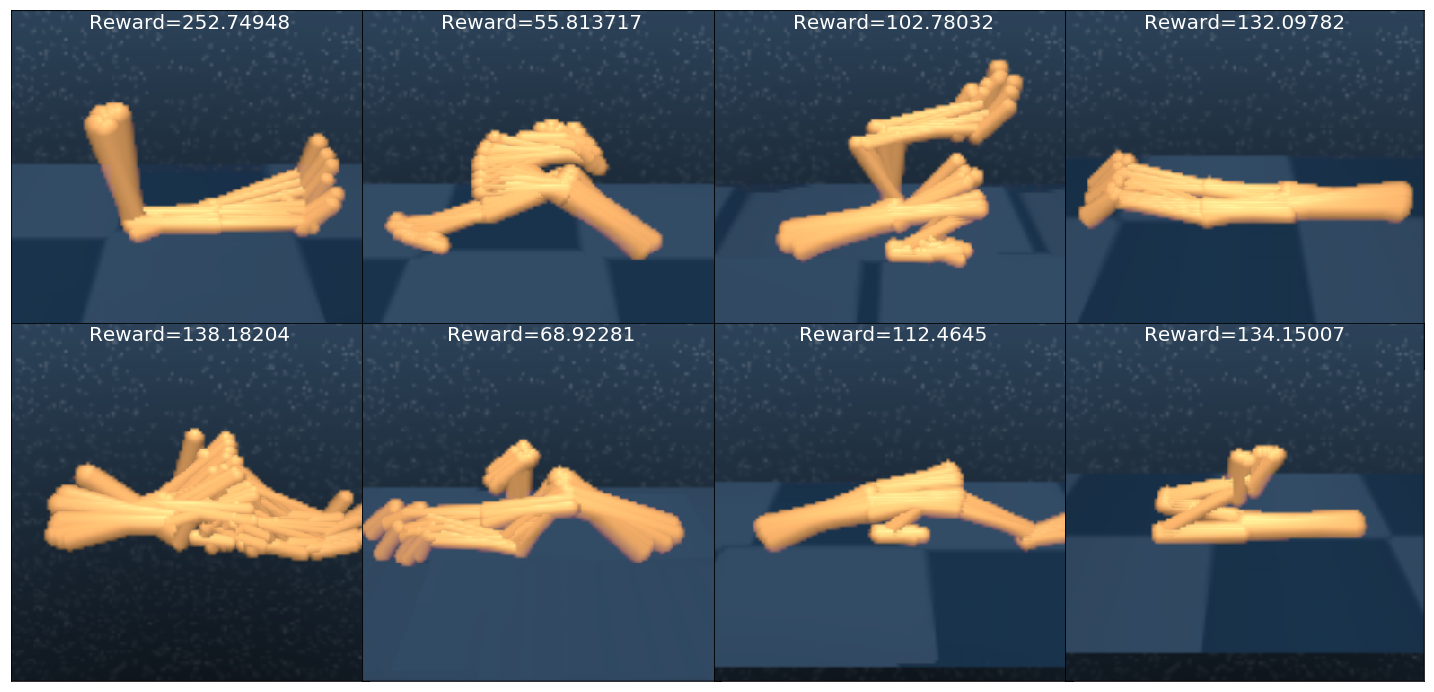}
\caption{Min}
\label{fig:supp_walker_walk_min_smp}
\end{figure}
\begin{figure}[h]
\centering
\includegraphics[width=0.9\linewidth]{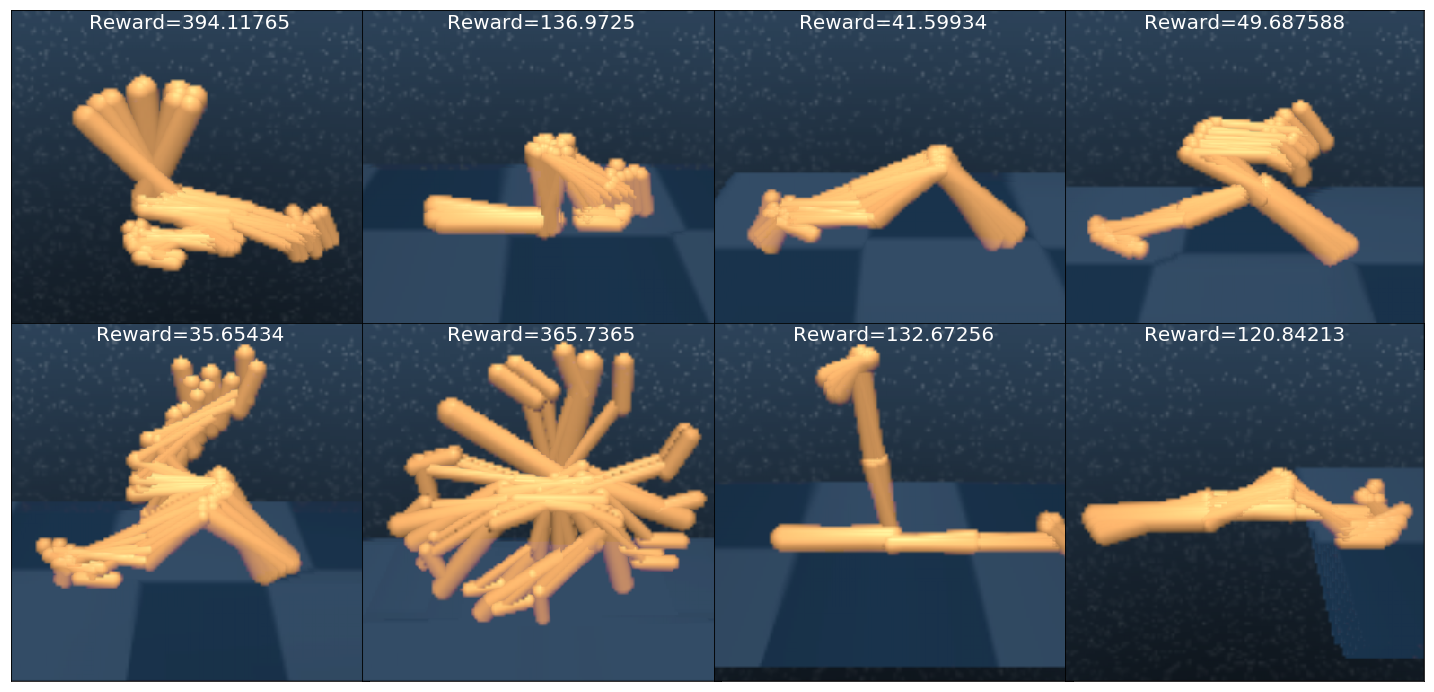}
\caption{Average}
\label{fig:supp_walker_walk_avg_smp}
\end{figure}
\begin{figure}[h]
\centering
\includegraphics[width=0.9\linewidth]{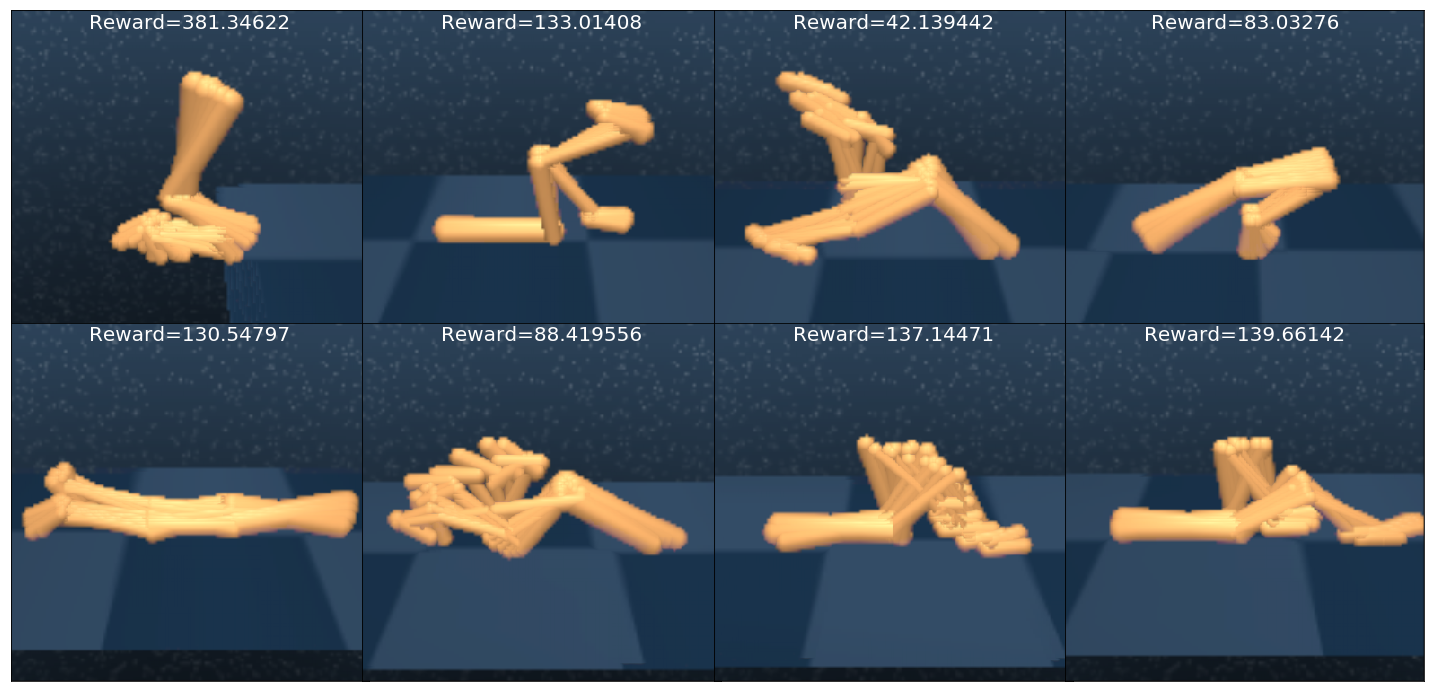}
\caption{Discrimination}
\label{fig:supp_walker_walk_disc_smp}
\end{figure}
\begin{figure}[h]
\centering
\includegraphics[width=0.9\linewidth]{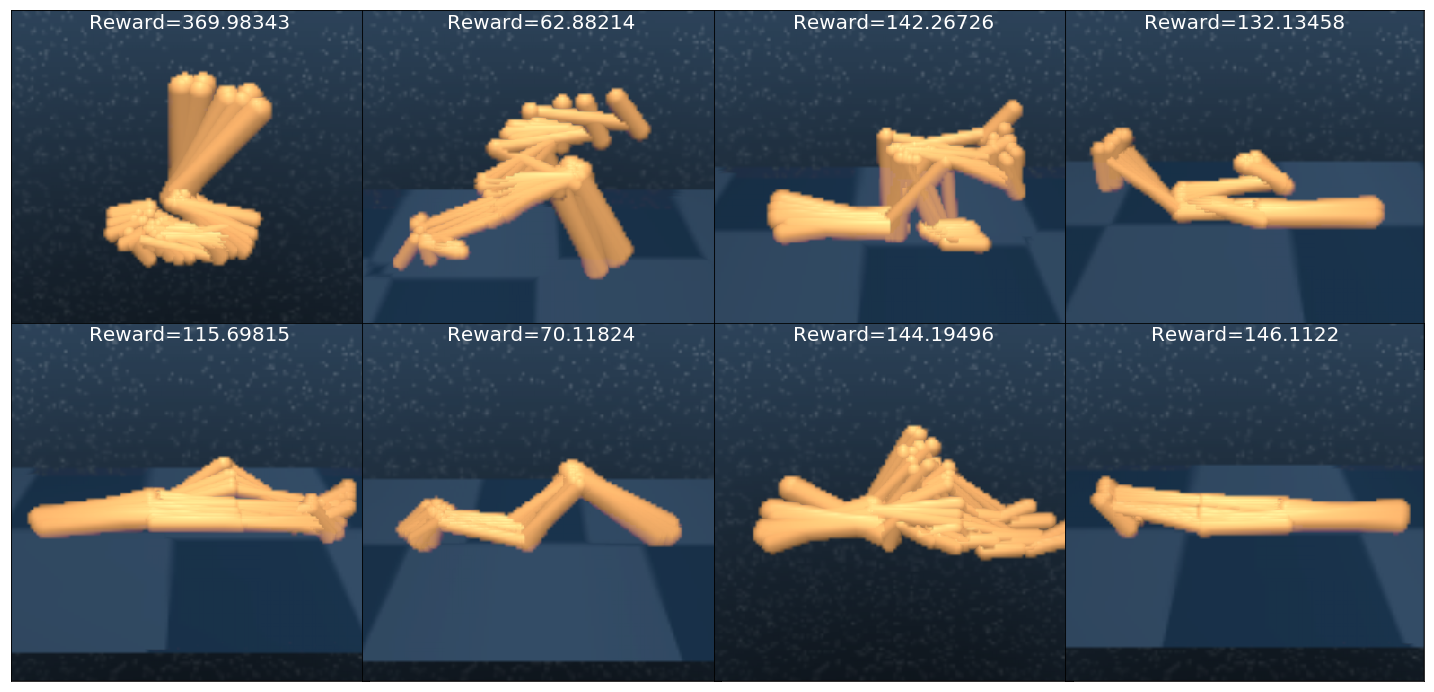}
\caption{None}
\label{fig:supp_walker_walk_none_smp}
\end{figure}

\clearpage
\subsection{Robustness in Cheetah}
\label{subsec:cheetah_smp}
\begin{figure}[h]
\centering
\includegraphics[width=0.9\linewidth]{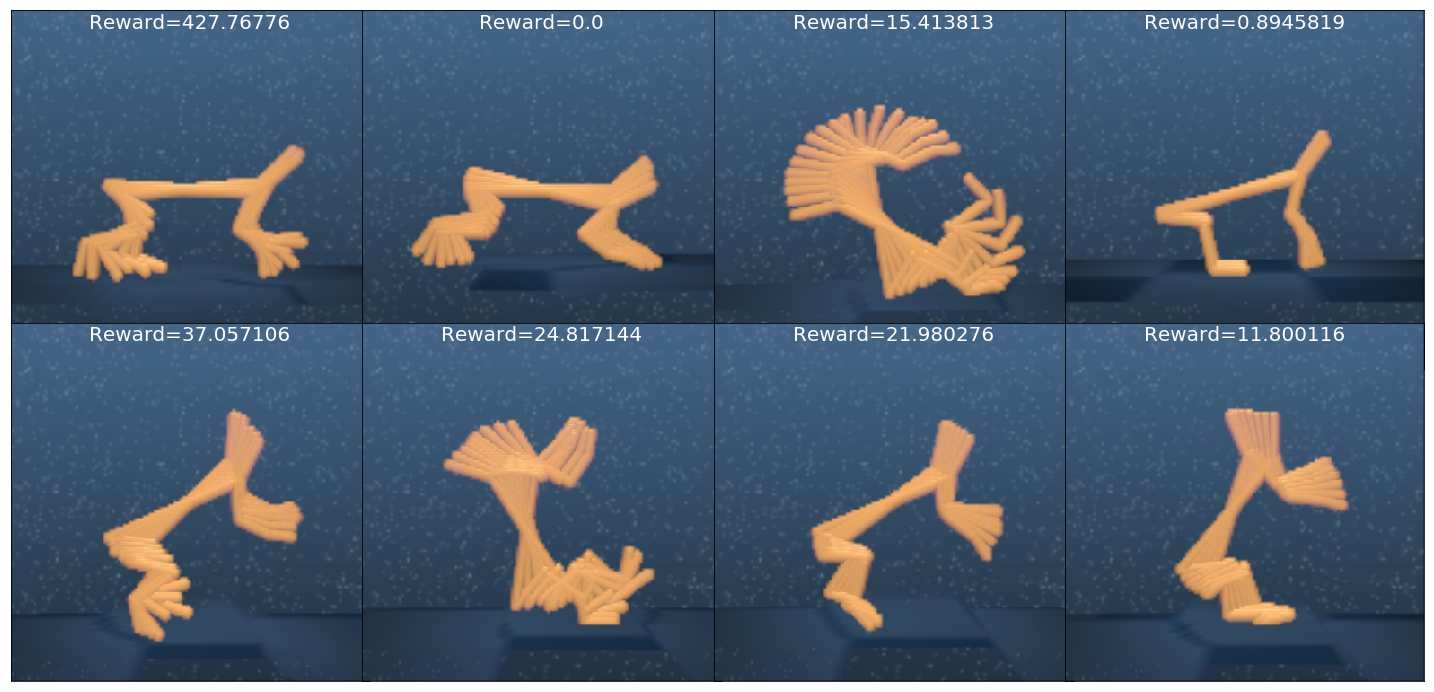}
\caption{Min}
\label{fig:supp_cheetah_walk_min_smp}
\end{figure}
\begin{figure}[h]
\centering
\includegraphics[width=0.9\linewidth]{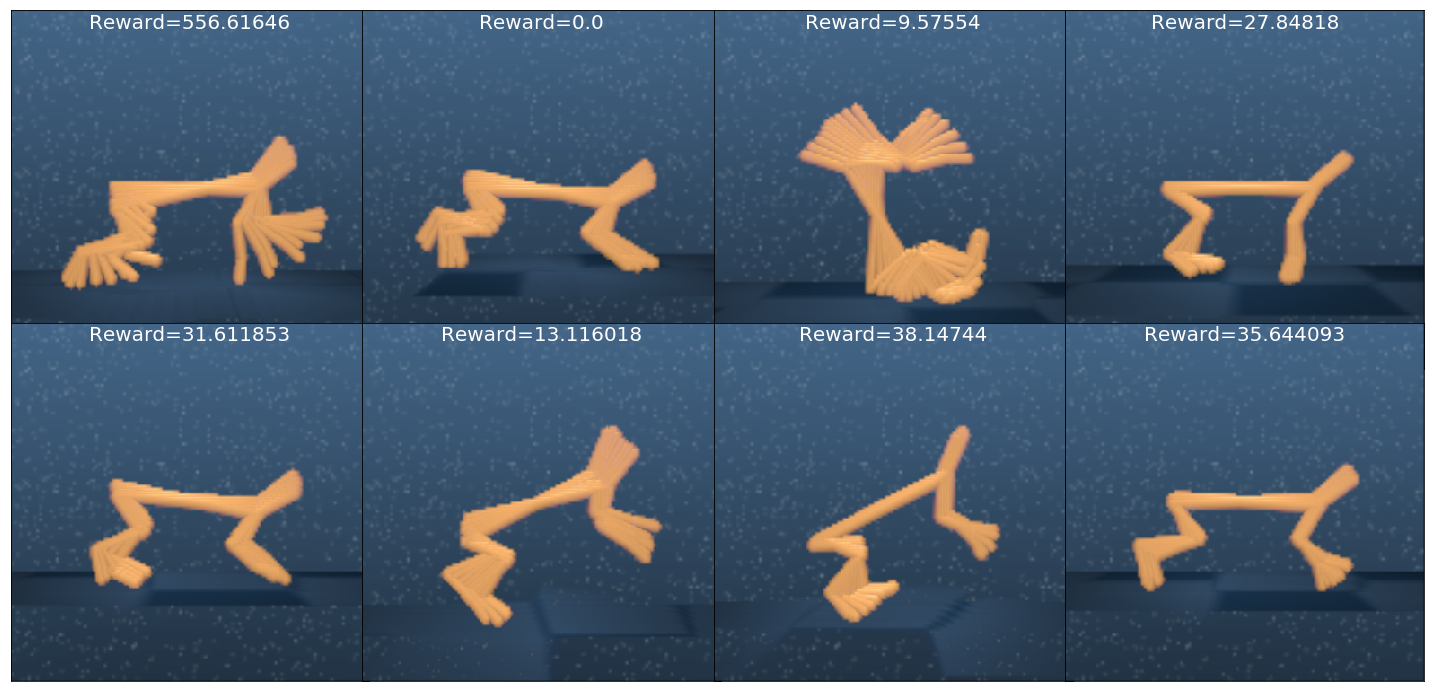}
\caption{Average}
\label{fig:supp_cheetah_walk_avg_smp}
\end{figure}
\begin{figure}[h]
\centering
\includegraphics[width=0.9\linewidth]{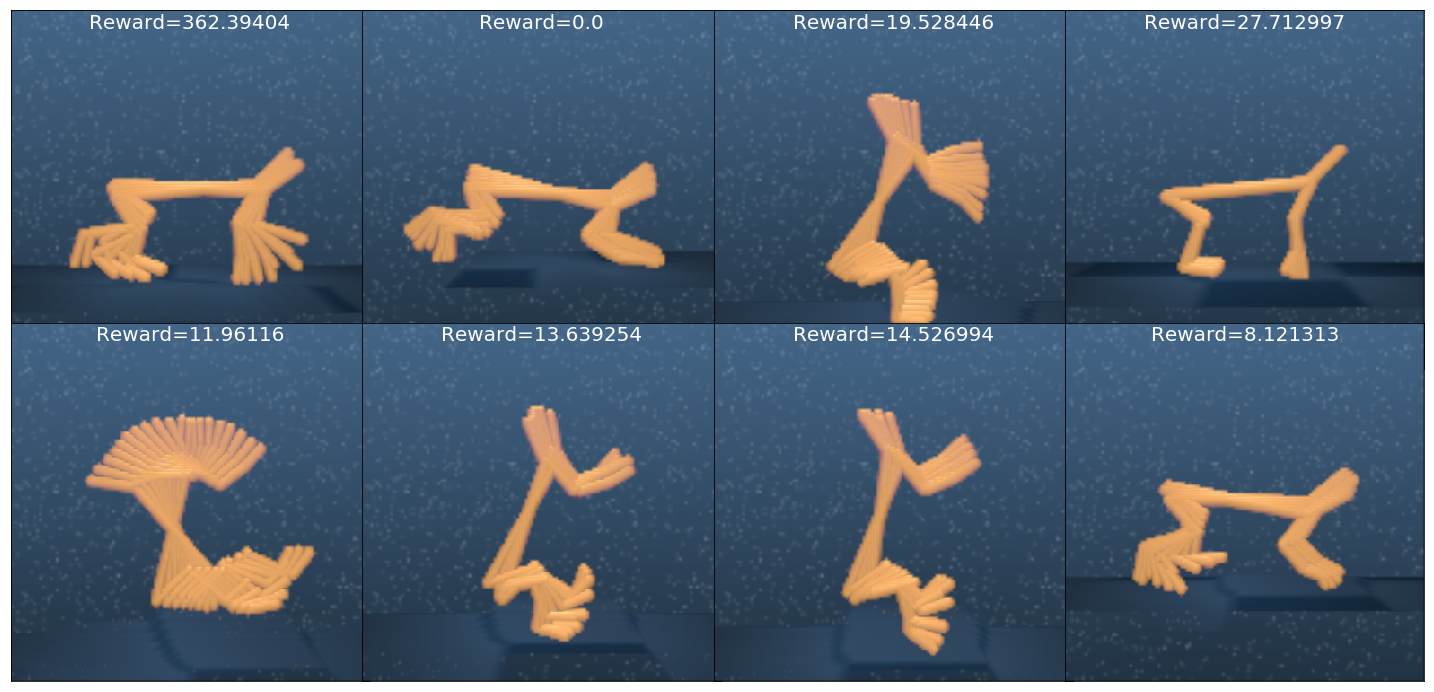}
\caption{Discrimination}
\label{fig:supp_cheetah_walk_disc_smp}
\end{figure}
\begin{figure}[h]
\centering
\includegraphics[width=0.9\linewidth]{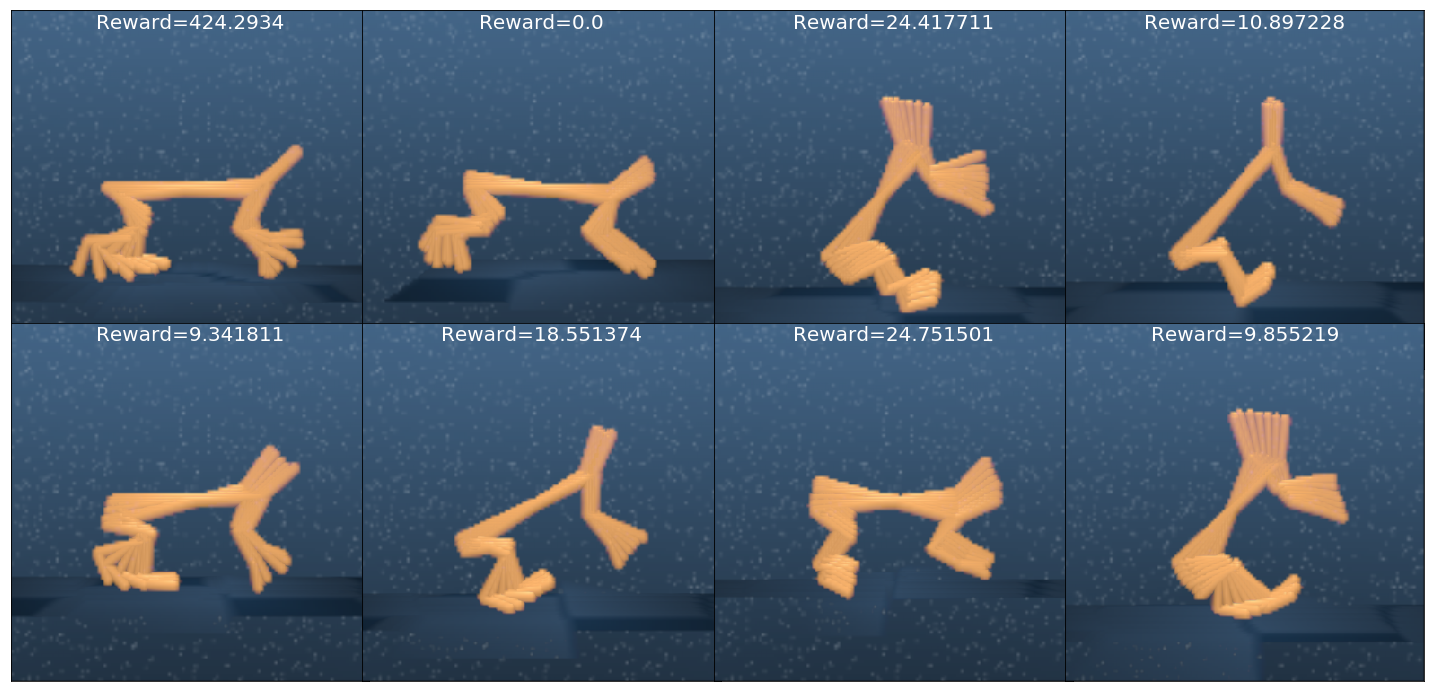}
\caption{None}
\label{fig:supp_cheetah_walk_none_smp}
\end{figure}

\end{document}